\documentclass[letterpaper]{article} 
\usepackage[draft]{aaai2027}  
\usepackage{times}  
\usepackage{helvet}  
\usepackage{courier}  
\usepackage[hyphens]{url}  
\usepackage{graphicx} 
\usepackage{natbib}  
\usepackage{caption} 
\usepackage{amsmath}
\usepackage{amssymb}
\usepackage{nicematrix}
\usepackage{bm}
\usepackage{xcolor}
\usepackage{subcaption}
\usepackage{booktabs}
\usepackage{tabularx}
\usepackage{array}
\usepackage{makecell}
\usepackage{amssymb}
\usepackage{mathtools}
\usepackage{algorithm}
\usepackage{algorithmic}
\usepackage{float}
\usepackage{fontawesome5}
\usepackage{newfloat}
\usepackage{listings}
\DeclareCaptionStyle{ruled}{labelfont=normalfont,labelsep=colon,strut=off} 
\floatstyle{ruled}
\newfloat{listing}{tb}{lst}{}
\floatname{listing}{Listing}
\usepackage{xspace}
\newcommand{\ours}[0]{{GAINS}\xspace}
\newcommand{\newbench}[0]{{HIL-RM}\xspace}

\title{GAINS: Leveraging Inconsistent Human Intervention Signals \\in Reinforcement Learning}
\author{
Xinyi Zhang\textsuperscript{\rm 1,2,$*$},
Yinuo Zhao\textsuperscript{\rm 3,2,$*$},
Pei Ren\textsuperscript{\rm 2,$\dagger$},
Lechun Jiang\textsuperscript{\rm 4,2},
Huiqian Jin\textsuperscript{\rm 4,2} \\
Lei Sun\textsuperscript{\rm 4},
Dapeng Wu\textsuperscript{\rm 3},
Zhengping Che\textsuperscript{\rm 2,$\dagger$},
Chi Harold Liu\textsuperscript{\rm 1},
Jian Tang\textsuperscript{\rm 2 \faIcon[regular]{envelope}}
}

\affiliations {
    \textsuperscript{\rm 1} Beijing Institute of Technology\\ 
    \textsuperscript{\rm 2} Beijing Innovation Center of Humanoid Robotics \\
    \textsuperscript{\rm 3} City University of Hong Kong \\
    \textsuperscript{\rm 4} Nankai University \\
    xinyizhang25@bit.edu.cn, \  yinuzhao@cityu.edu.hk, \  \{pei.ren,z.che,jian.tang\}@x-humanoid.com
}

\usepackage{bibentry}

\begin{document}

\maketitle

\begingroup
\renewcommand{\thefootnote}{}
\footnotetext{%
\raggedright
\textsuperscript{*}\ Equal contribution: Xinyi Zhang and Yinuo Zhao.\\
\faIcon[regular]{envelope}\ Corresponding author: Jian Tang
.
\textsuperscript{$\dagger$}\ Project leaders: Zhengping Che and Pei Ren.
This work was conducted during Xinyi Zhang's internship at the Beijing
Innovation Center of Humanoid Robotics.
}
\endgroup
\begin{abstract}
Correcting robot manipulation policies through human intervention holds great promise for real-world deployment, yet human operators are inherently imperfect in both the actions they provide and the timing of their intervention signals. While the former has been extensively discussed in reinforcement learning (RL), the latter remains underexplored. At high control frequencies, human intervention signals are often delayed and inconsistent across time and state space. In this work, we present \textbf{\ours}, a framework for 
levera\textbf{G}ing inconsistent hum\textbf{A}n \textbf{I}nterventio\textbf{N} \textbf{S}ignals 
in RL. 
At the core of \textbf{\ours}, we employ distributional RL with quantile Q-networks to model the return variability induced by sparse task rewards and inconsistent human interventions.
Building on this distributional representation, we introduce a pessimistic exploration strategy that promotes safe and sample-efficient learning under human corrections.
We evaluate \textbf{\ours} on four diverse simulated manipulation tasks and two challenging real-world scenarios against state-of-the-art intervention-based methods. \textbf{\ours} achieves a 22\% higher task success rate than RLIF and improves recovery success by up to 43\% in failure scenarios. These results highlight the importance of modeling return variability induced by human imperfection for real-world deployment of intervention-based learning.
Project website: 
https://gains-hil.github.io/.
\end{abstract}

\section{Introduction}
Learning robot manipulation policies through human intervention has emerged as a promising paradigm for safe and efficient real-world deployment~\cite{luo2025precise,chen2025conrft}. Unlike purely offline systems that struggle with distribution shifts and unforeseen failures, human-in-the-loop (HIL) frameworks allow human operators to provide corrective feedback when the robot behaves suboptimally. Such feedback enables rapid online learning without requiring extensive offline demonstrations. Moreover, as model capabilities improve, the required human effort decreases, making HIL a practical approach for complex manipulation tasks.

Existing HIL methods can be broadly categorized by how they use human feedback. The first category, \emph{learning from intervention actions}, directly imitates corrective actions provided by human operators~\cite{kelly2019hg}. Although intuitive, these methods are inherently limited by the quality and consistency of human actions, particularly in tasks that require greater precision. The second category, \emph{learning from intervention signals}~\cite{luo2024rlif,korkmaz2025mile}, treats intervention events as binary indicators of policy failure and uses them as feedback for RL-based policy improvement. By relying on intervention signals without assuming optimal corrective actions, these methods can improve the policy beyond the quality of the human actions.

\begin{figure}
    \centering
    \includegraphics[width=\linewidth]{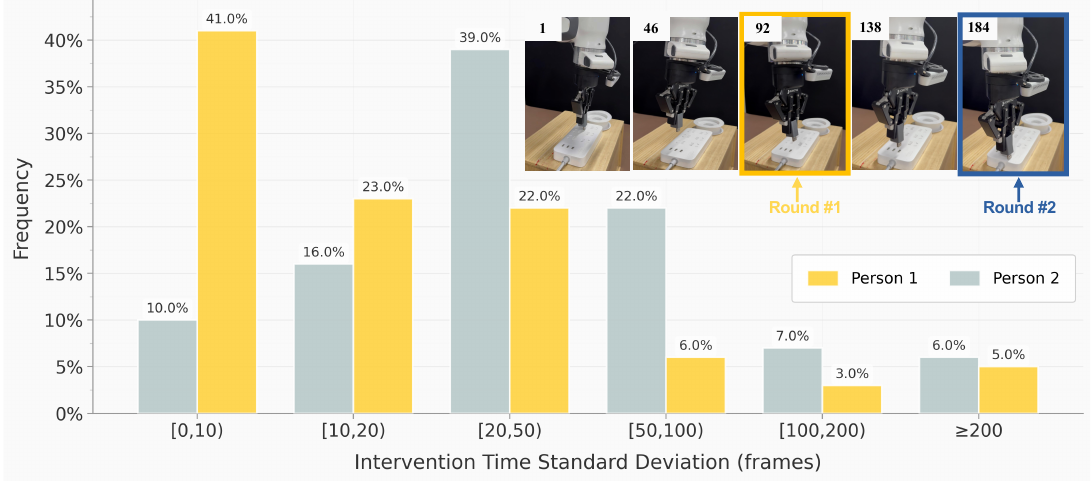}
    \caption{
    \textbf{Human Intervention Timing Is Inconsistent.} We replay failure videos multiple times to measure intervention-timing consistency; the top-right panel highlights intervention frames from different rounds in yellow.}
    \label{fig:intuition}
\end{figure}

However, existing signal-based methods, such as RLIF~\cite{luo2024rlif}, treat each intervention as a precise binary label and model them with scalar Q-functions. This formulation does not explicitly account for uncertainty or inconsistency in human intervention signals. To study this issue, we collected 100 failure videos from 52 manipulation tasks performed by $\pi_0$~\cite{black2024pi_0}. Two operators watched each video five times and marked when they would intervene. As shown in Fig.~\ref{fig:intuition}, both operators exhibit substantial inconsistency, with over 50\% of interventions differing in timing by more than 10 frames at 30 fps. The second operator (gray) shows greater variability: 39\% differ by 20--50 frames, and 13\% by more than 100 frames, corresponding to over 3.3 seconds. These findings demonstrate that human intervention signals vary not only across repeated observations but also across operators and state regions. Consequently, the return distribution induced by human interventions can be broad or multimodal, which cannot be fully characterized by a scalar estimate of the expected return.

Distributional RL models uncertainty by estimating the full return distribution rather than only its expectation~\cite{dabney2018distributional,bellemare2017distributional}. Quantile-based value estimation can capture multimodal returns and support risk-sensitive policy optimization. Despite these advantages, distributional RL remains largely unexplored in HIL settings. In particular, it remains unclear how distributional value representations can be used to safely and efficiently learn from inconsistent intervention signals.

To address this challenge, we propose \textbf{\ours}, a framework for levera\textbf{G}ing inconsistent hum\textbf{A}n \textbf{I}nterventio\textbf{N} \textbf{S}ignals in reinforcement learning. \textbf{\ours} uses truncated distributional Q-estimation to model the mixed returns caused by inconsistent interventions and sparse task rewards. Based on this distribution, we introduce a reward-scaled quantile Huber loss for stable value learning and a pessimistic exploration strategy that favors actions less likely to trigger human intervention. These components are integrated into an asynchronous actor--critic framework for safe and sample-efficient online learning.

Existing HIL methods are often evaluated either in real-world or relatively simple simulated environments~\cite{hu2022model}, making systematic and reproducible comparisons difficult. To address this limitation, we introduce \textbf{\newbench}, an HIL robot manipulation training suite with four tasks and support for flexible human intervention. We evaluate \textbf{\ours} against intervention-based baselines on \textbf{\newbench} and further validate them on two real-world manipulation tasks. Our main contributions are summarized as follows:
\begin{itemize}
\item We identify and empirically characterize the inconsistency of human intervention signals. Based on this observation, we propose \textbf{\ours}, a distributional HIL-RL framework that explicitly models the return variability induced by inconsistent interventions.

\item Based on a truncated distributional Q-estimation, we further propose a reward-scaled quantile Huber loss, and a pessimistic exploration strategy within an asynchronous actor--critic framework for efficient online learning.

\item We introduce \textbf{\newbench}, an HIL robot manipulation training suite with four simulated tasks and support for human intervention. On \textbf{\newbench}, \textbf{\ours} improves recovery success by 11\%--43\% over the evaluated baselines across failure scenarios. On two challenging real-world tasks, \textbf{\ours} improves task success by up to 52.5\%.
\end{itemize}

\section{Related Works}
\subsection{Human-in-the-loop Reinforcement Learning}
Training vision-based robot manipulation policies directly in the real world remains challenging due to slow, risky, and expensive exploration. 

HIL-SERL~\cite{luo2025precise} achieves superhuman performance by combining sample-efficient off-policy RL with a pretrained visual backbone. ConRFT~\cite{chen2025conrft} further adopts a larger pretrained vision-language encoder and uses a consistent policy across offline and online training. Despite its strong performance, ConRFT retains an imitation loss during online training, making it more dependent on the quality of human demonstrations. AWAC~\cite{nair2020awac} shares this dependency, bridging offline and online learning via advantage-weighted behavior cloning.

Another line of work leverages human intervention signals during online RL, in both robotics~\cite{luo2024rlif} and autonomous driving~\cite{huang2024human,peng2022safe,peng2023learning}. However, these methods assume reliable human feedback, whereas real-world interventions are often delayed, inconsistent, or noisy. In this work, we instead focus on learning from imperfect human interventions under sparse task rewards, rather than filtering out inconsistent samples or assuming interventions arise from deterministic conditions.

\begin{figure*}[htb]
    \centering
    \includegraphics[width=\linewidth]{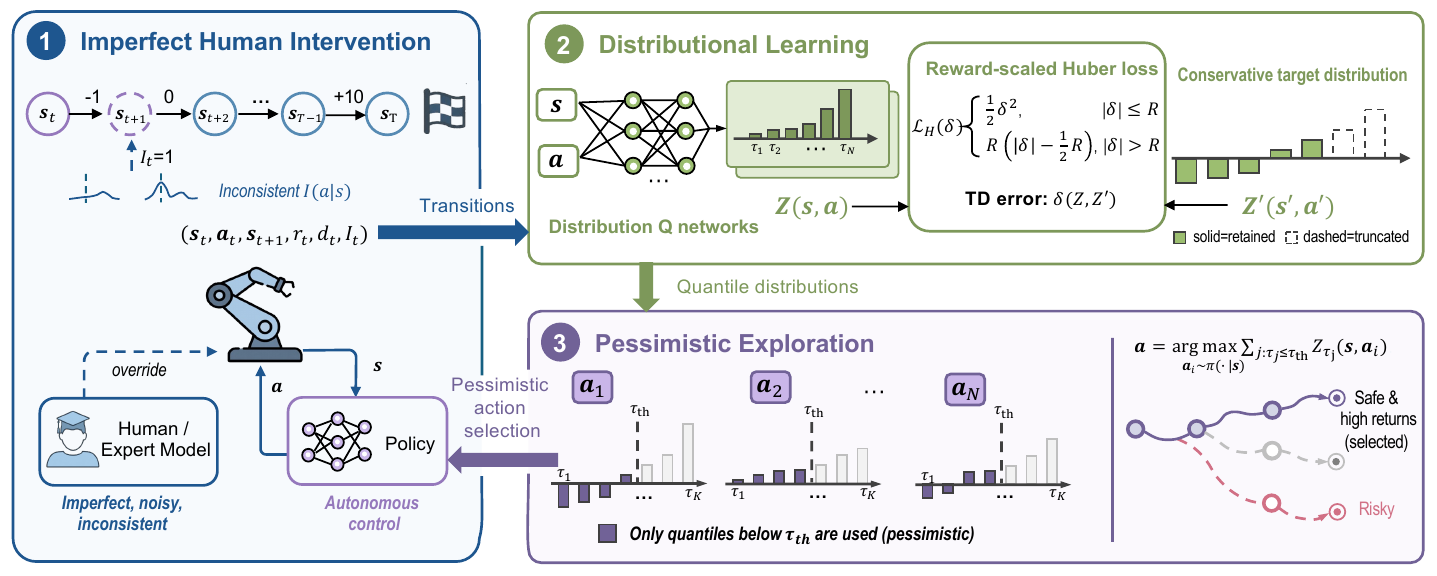}
    \caption{An Overview of \ours.}
    \label{fig:overview}
\end{figure*}

\subsection{Interactive Imitation Learning (IL)}
Interactive IL is another paradigm for online policy learning, categorized by the domain of human feedback~\cite{celemin2022interactive}. In state--action domain, humans demonstrate or correct robot actions~\cite{ross2011reduction,macglashan2017interactive,kelly2019hg,mandlekar2020human}, typically assuming that such feedback is optimal. In evaluative domain, humans provide scalar feedback on the quality of the agent's behavior~\cite{knox2009interactively,christiano2017deep,ibarz2018reward}. Under this taxonomy, HIL-RL methods can also be viewed as evaluative approaches. Among these, Sirius~\cite{conf/rss/LiuNZBZ23} is a representative method that uses human intervention signals to estimate human trust and reweight training samples accordingly.

\subsection{Distributional RL}
\paragraph{Quantile Regression} In distributional RL, the return is modeled as a distribution instead of a scalar value.  The distributional Bellman operator is defined as:
\begin{equation}
    \mathcal{T}^\pi Z(\bm{s}, \bm{a}) \stackrel{D}{=} R(\bm{s}, \bm{a}) + \gamma Z(\bm{s}', \bm{a}'),
\end{equation}
where $\bm{s}' \sim P(\cdot|\bm{s}, \bm{a})$ and $\bm{a}' \sim \pi(\cdot|\bm{s}')$. Early works~\cite{bellemare2017distributional,dabney2018distributional,dabney2018implicit} established the theoretical foundations of distributional RL, primarily for discrete action spaces. TQC~\cite{kuznetsov2020controlling} extends it to continuous control by combining distributional critics with truncated value estimation under SAC~\cite{haarnoja2018soft}.

\paragraph{Pessimism in Offline and Safe RL}
Pessimistic value estimation is widely used to reduce Q-value overestimation and unsafe exploration. In offline RL, CQL~\cite{kumar2020conservative} penalizes high Q-values for out-of-distribution actions. Ensemble-based methods such as PBRL~\cite{DBLP:conf/rss/HuMS24} select between actions proposed by imitation and RL policies to avoid unreliable RL actions. In distributional RL, IQN~\cite{dabney2018implicit} enables risk-sensitive control by optimizing distorted risk measures, such as CVaR, over the learned return distribution. Our method follows this distributional perspective but uses lower-tail return estimates during action selection to avoid states likely to trigger human intervention. This differs from prior CVaR-based methods, which mainly target general risk-sensitive control or predefined safety constraints.

\section{Preliminary}

TQC~\cite{kuznetsov2020controlling} is a representative distributional RL method built on SAC~\cite{haarnoja2018soft} to reduce overestimation bias in Q-value estimation. Instead of learning a scalar Q-function, TQC maintains $M$ critic networks parameterized by ${\theta}_{m=1}^{M}$. Given a state-action pair $(s,a)$, the $m$-th critic outputs $K$ scalar quantile atoms:
\begin{equation}
    \left\{ z_i^m(s,a;\theta) \right\}_{i=1}^{K},
    \quad m = 1,\dots,M,
\end{equation}
where $z_i^m$ denotes the $\tau_i$-th quantile of the $m$-th critic.

To construct the distributional Bellman target, TQC first samples the next action 
$a' \sim \pi(\cdot \mid s')$ and evaluates it using the target critics. Specifically, all $M \times K$ target atoms are collected as
\begin{equation}
\hat{\mathcal{Z}}(s',a') =
\left\{
\hat{z}_i^m(s',a')
\right\}_{m=1,i=1}^{M,K}.
\end{equation}
These atoms are sorted in ascending order, and the largest $d$ atoms are truncated to suppress overly optimistic return estimates:
\begin{equation}
\left\{
\hat{z}_j(s',a')
\right\}_{j=1}^{M K-d}
=
\operatorname{sort}\!\left(
\hat{\mathcal{Z}}(s',a')
\right)_{1:M K-d}.
\label{eqn:tqc_Q}
\end{equation}
The truncated Bellman target atoms are then formed as $y_j = r + \gamma \hat{z}_j(s',a')$ for $j=1, \cdots, MK-d$. We omit the terminal mask and the SAC entropy term in the Bellman target for clarity. The temporal difference error are computed as:
\begin{equation}
    \delta_{i,j,m} = y_j-z_i^m(s,a;\theta).
    \label{eqn:td_error}
\end{equation}
And the critics are trained by minimizing the quantile Huber loss between current quantile and truncated target atoms:
\begin{equation}
\mathcal{L}_{critic}(\theta)=
\tfrac{1}{MK(MK{-}d)}
\sum_{m=1}^{M}\!\sum_{i=1}^{K}\!\sum_{j=1}^{MK{-}d}
\mathcal{L}_{\tau_i}\left(\delta_{i,j,m}\right),
\label{eqn:critic_loss}
\end{equation}
where $\mathcal{L}_{\tau_i}(\cdot)$ is the quantile Huber loss discussed later.

We observe that Q-value overestimation can happen in HIL-RL, especially during real-world training, where the robot frequently revisits undesired states even after repeated human corrections. This issue stems from several compounding factors: the distribution discrepancy between human interventions and the learned policy, the low data collection speed relative to the number of gradient update iterations, function approximation error, and the maximization bias introduced by the max operator in standard actor-critic methods. TQC's truncation mechanism mitigates this overestimation by discarding upper-tail quantile atoms from the return distribution, making it a principled backbone for HIL-RL. Building on this foundation, we present our method, which introduces targeted contributions to address the practical challenges of sparse reward settings in HIL-RL.

\section{Method}
In this section, we present \ours, a framework for robust learning from imperfect human corrections in human-in-the-loop reinforcement learning (HIL-RL). Fig.~\ref{fig:overview} provides an overview of our method. \ours treats the return induced by intervention signals and binary environment rewards as inherently multi-modal, and therefore estimates the full return distribution rather than a scalar expectation. We first formalize HIL-RL as a Markov decision process. We then describe the distributional Q-learning and policy optimization procedures at the core of \ours, and introduce a pessimistic exploration strategy that exploits the learned return distribution. Implementation details are provided in Appendix~\ref{supp:gains_details}.

\subsection{Problem Formulation}

As shown in the left panel of Fig.~\ref{fig:overview}, we consider an HIL-RL setting in which a human can take control from the robot policy to support safe and efficient online training. We account for two forms of imperfect human feedback. First, intervention actions may be suboptimal, as commonly considered in prior HIL-RL work. Second, intervention signals may be inconsistent: due to high control frequencies, continuous transition spaces, and human fatigue. Noted that we neither inject artificial noise into these signals nor require an explicit estimate of their inconsistency.

We formulate robot manipulation as an MDP defined by the tuple $(\mathcal{S},\mathcal{A},\mathcal{I},\mathcal{P},\mathcal{R})$. Each state $\bm{s}\in\mathcal{S}$ consists of RGB images and robot proprioception. For simplicity, we refer to $\bm{s}$ as the state, although it is an observation that only partially captures the underlying environment state. An action $\bm{a}\in\mathcal{A}$, such as a desired 6-DoF end-effector pose increment and a gripper command, is generated either by the learned policy $\pi$ or by a human operator through teleoperation. The intervention indicator $I_t\in\mathcal{I}={0,1}$ specifies the action source, where $I_t=1$ denotes human control and $I_t=0$ denotes autonomous control. $\mathcal{P}$ denotes the transition dynamics, and $\mathcal{R}:\mathcal{S}\times\mathcal{A}\rightarrow\mathbb{R}$ denotes the reward function.

For all tasks, the agent receives a reward of $+10$ for task success and a penalty of $-1$ for a safety violation or for the transition immediately before an intervention. All other transitions receive a reward of $0$ in simulation or a time penalty of $-0.05$ in real-world experiments. In practice, sparse success rewards can be provided by a binary success classifier~\cite{luo2025precise,ball2023efficient,chen2025conrft} or directly by a human operator~\cite{li2025gr}. The policy objective is
\begin{equation}
\max_{\pi}\quad
J(\pi)
\coloneqq
\mathbb{E}{\rho^{\pi}}
\left[
\sum{t=0}^{\infty}\gamma^{t}
r(\bm{s}_t,\bm{a}_t)
\right],
\label{eqn}
\end{equation}
where $\rho^\pi$ denotes the state--action visitation distribution induced by autonomous rollouts and human interventions.

\paragraph{Return variability under inconsistent interventions.}
We model the human intervention signal as a potentially stochastic policy $I_t \sim \mu_H(\cdot \mid h_t)$, where $h_t$ contains the current observation and interaction history. Due to reaction delay or subjective judgment, similar states may receive interventions at different time. Consequently, even for similar state--action pairs, different intervention sequences can lead to different intervention penalties and task outcomes. The resulting return distribution is as:
$
p(Z\mid s,a)
=
\sum_{\xi}
p(\xi\mid s,a)p(Z\mid s,a,\xi),    
$
where $\xi$ denotes a possible future intervention sequence. A scalar
critic collapses this heterogeneous distribution into its expectation,
whereas a distributional critic retains its lower-tail and multimodal
structure. GAINS does not explicitly identify the intervention-delay
distribution; instead, it learns a return distribution that is robust
to the outcome variability induced in part by inconsistent
interventions.

\subsection{Distributional Learning}

\paragraph{Critic Update} 
As discussed above, the return distribution captures the mixed effects of sparse task rewards, human intervention signals, and safety-violation penalties (the latter applied only in simulation, since safety violations are not allowed in practice). To model this distribution, \ours adopts truncated distributional Q-estimation, which discards the largest $d$ target atoms when constructing the target return distribution, as defined in Eq.~\eqref{eqn:tqc_Q}. Then an asymmetric quantile Huber loss is applied to compute the critic loss while reducing the effect of outliers:
\begin{equation}
    \mathcal{L}_{\tau}(\delta) = \left|\tau - \mathbf{1}[\delta < 0]\right| \cdot \mathcal{L}_H(\delta), 
\end{equation}
where $\delta$ denotes the critic prediction error under each quantile, and 
\begin{equation}
\mathcal{L}_{H}(\delta)
=
\begin{cases}
\dfrac{1}{2}\delta^2, & |\delta| \leq \kappa, \\[6pt]
\kappa\left(|\delta| - \dfrac{\kappa}{2}\right), & |\delta| > \kappa.
\end{cases}
\end{equation}

In standard distributional RL, the Huber threshold $\kappa$ is typically set to $1$, which is suitable for dense rewards with unit-scale magnitudes. However, this choice may be unsuitable in sparse-reward tasks, where success rewards are large but rare. Near successful transitions, the TD error $\delta$ can increase sharply, causing a small threshold to move informative errors into the linear regime too early and weaken reward propagation. We therefore set $\kappa=R$, where $R$ is the success reward. Errors within $[-R,R]$ are penalized quadratically, providing smooth and strong gradients for value learning. Larger errors are penalized linearly, limiting excessive updates caused by severe Q-value overestimation. We refer to this formulation as the \emph{reward-scaled quantile Huber loss}. The overall critic loss averages this loss across all critics and predicted--target quantile pairs, as defined in Eq.~\eqref{eqn:critic_loss}.

\paragraph{Actor Update}
The actor is optimized with the entropy-regularized objective:
\begin{equation}
    \mathcal{L}_{\mathrm{actor}}(\phi)
    =
    \mathbb{E}_{\bm{s}}
    \left[
        \alpha \log \pi_{\phi}(\bm{a}_{\pi}|\bm{s})
        -
        Q_{\min}(\bm{s},\bm{a}_{\pi})
    \right],
\end{equation}
where $\bm{a}_{\pi}\sim\pi_{\phi}(\cdot\mid\bm{s})$ and $\alpha$ is the entropy temperature. $Q_{\min}$ denotes the minimum scalar $Q$ value across the critic ensemble:
$    Q_{\min}(\bm{s},a_{\pi})
    =
    \min_{m=1,\dots,M}
    Q^{m}(\bm{s},a_{\pi}).$

For each critic, the scalar Q-value is computed by averaging its $K$ quantile atoms:
$Q^{m}(\bm{s},\bm{a}_{\pi})=\frac{1}{K}\sum_{k=1}^{K}z_{k}^{m}(\bm{s},\bm{a}_{\pi}).$

\paragraph{Discrete Critic Update}
Following HIL-SERL~\cite{luo2025precise}, we use separate models for continuous end-effector control and discrete gripper control. Because the gripper operates at a lower frequency, we model it with vanilla Double DQN~\cite{van2016deep} over a binary action space $a\in{0,1}$. In addition to the sparse task reward and intervention penalty, we apply a penalty of $-0.05$ whenever the gripper state changes. The discrete critic is then optimized via the standard Double DQN loss~\cite{van2016deep}:
\begin{equation}
\mathcal{L}(\theta)
\;=\;
\mathbb{E}_{\rho^\pi}
\Bigl[
\bigl(
Q(\bm{s},a;\theta) - y(\bm{s},a,r,\bm{s}')
\bigr)^2
\Bigr].
\label{eqn:ddqn_q_loss}
\end{equation}

\subsection{Pessimistic Exploration}

Even with truncated Q estimation, stochastic action sampling may still lead the agent to unsafe or inefficient states. Model-based RL~\cite{HansenSW22,TD-MPC2,DBLP:conf/iclr/HafnerLB020,gao2026dreamdojo} reduces this risk by evaluating candidate actions with a learned environment model before execution, but requires training an additional model. In contrast, distributional RL directly estimates the return distribution of each action, which can be used to guide action selection. To make the agent pay more attention to possible intervention and safety violation signals, we propose a pessimistic exploration strategy. Specifically, as shown in the third part of Fig.~\ref{fig:overview}, we first sample $N$ candidate actions from the policy $\pi(\cdot\mid\bm{s}_t)$. Then, we evaluate each candidate action by computing the mean of its lower-tail quantiles, defined by a threshold $\tau_{\mathrm{th}}$
its pessimistic Q estimate, defined as the average return over quantiles below a predefined threshold $\tau_\textrm{th}$, and execute the action with the highest pessimistic value:

\begin{equation}
\bm{a}_t = \arg\max_{\bm{a}_i \sim \pi(\cdot \mid \bm{s}_t)}
\sum_{\substack{j \in \{1,\dots,M\} \\ \tau_j \leq \tau_{\mathrm{th}}}} z_{\tau_j}(\bm{s}_t, \bm{a}_i)
\end{equation}
This strategy favors actions with better lower-tail outcomes and therefore reduces the risk of selecting actions that may trigger intervention or safety violations.

\section{Results}

In this section, we aim to address the following four research questions (\textbf{RQ}):

\noindent\textbf{RQ1:} How does \ours compare with existing baselines under human intervention?

\noindent\textbf{RQ2:} How robust are the policies learned by different methods under external disturbances?

\noindent\textbf{RQ3:} How does each component contribute to \ours?

\noindent\textbf{RQ4:} Can the distributional critic effectively capture the uncertainty induced by inconsistent human interventions?

Additional results, including an analysis of the quantile threshold $\tau_{\mathrm{th}}$, are provided in Appendix~\ref{supp:exp}.

\begin{figure}[tp]
    \centering
    \includegraphics[width=\columnwidth]{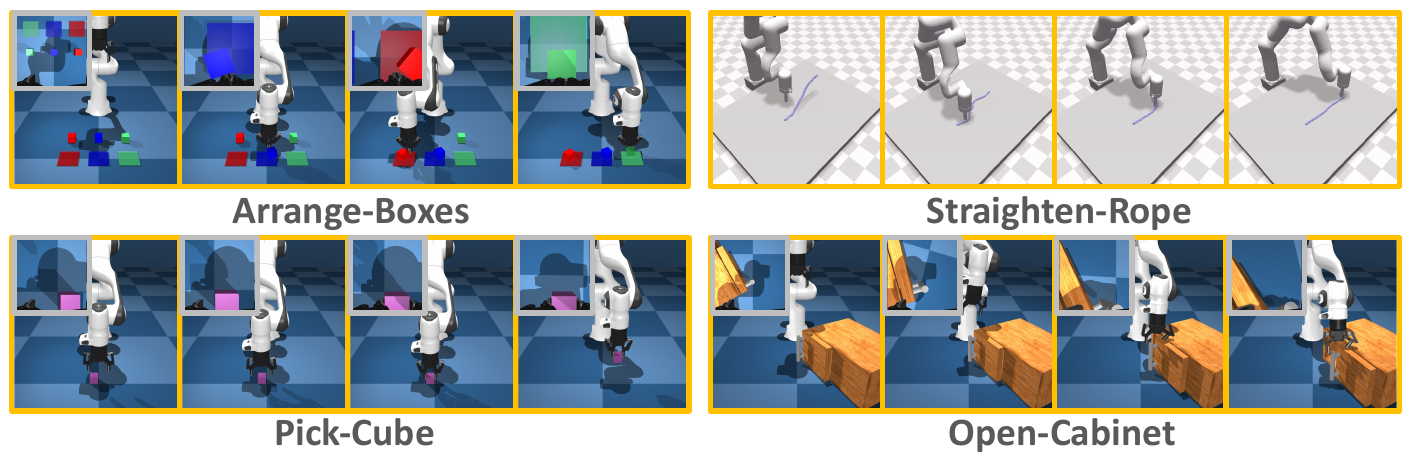}
    \caption{\textbf{Four Simulation Tasks in \newbench.} The top-left inset shows the wrist-camera view used as input for all tasks except Straighten-Rope, which uses a third-person camera.}
    \label{fig:sim_env}
\end{figure}

\begin{figure}[tp]
    \centering
    \includegraphics[width=\linewidth]{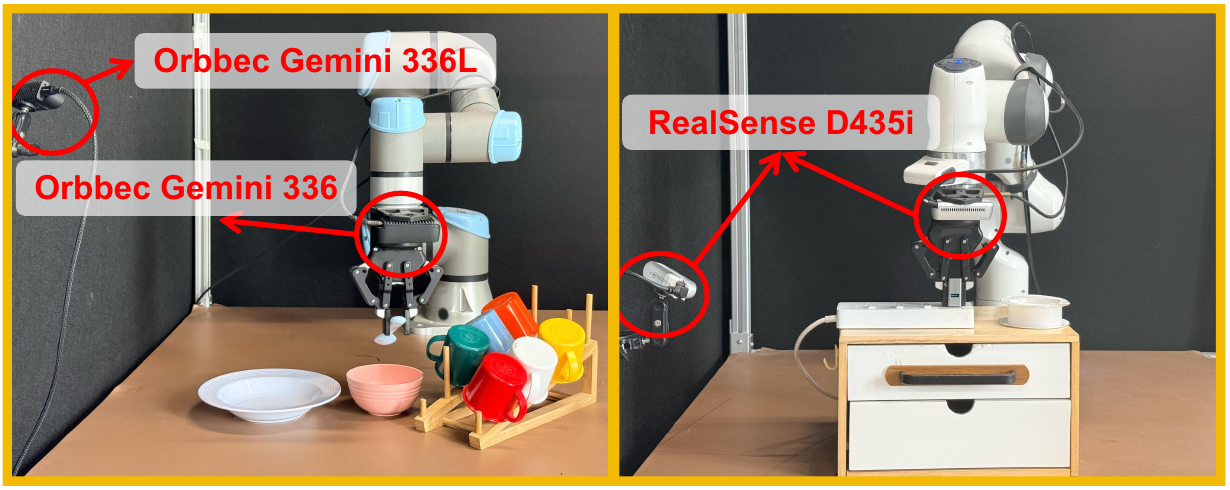}
    \caption{\textbf{Real-world Experimental Setups.} Left: UR5 with 2 cameras for the Pick-Spoon task. Right: Franka Emika with 2 cameras for the Insert-USB task.}
    \label{fig:real_world_env}
\end{figure}

\begin{figure*}[ht]
    \centering
    \includegraphics[width=\linewidth]{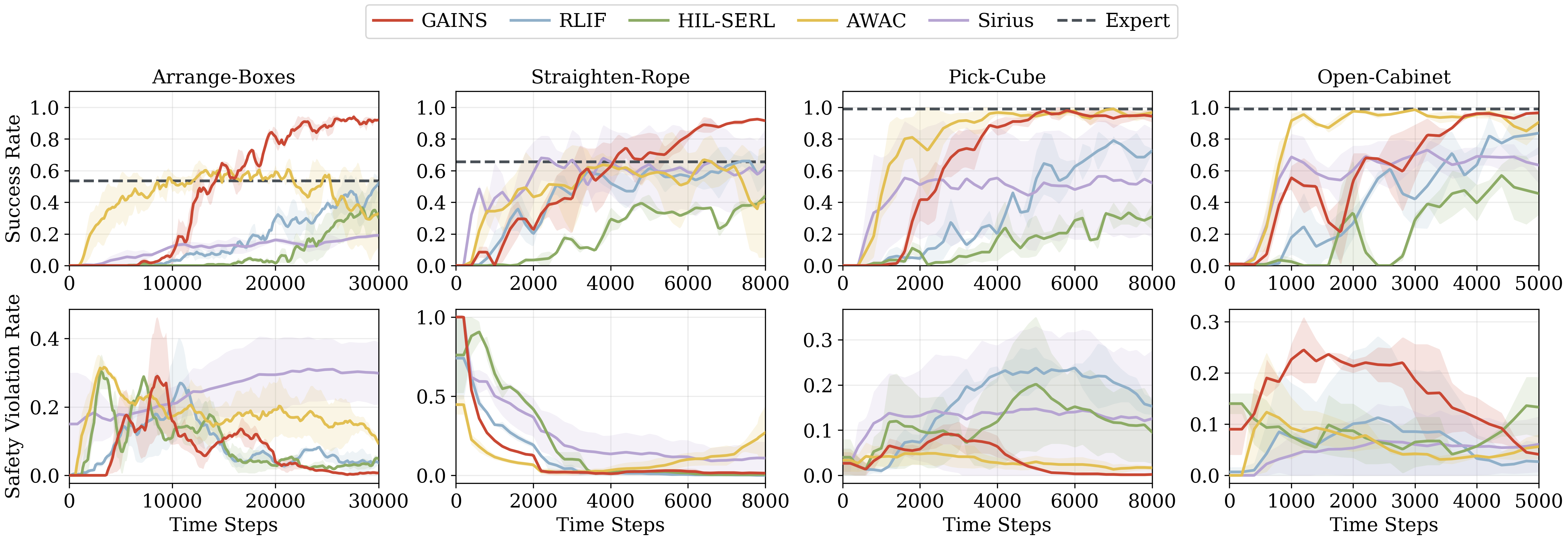}
    \caption{\textbf{Comparison Results in \newbench.} Results are averaged over 3 seeds, with shaded regions indicating the standard deviation. 
    Curves are smoothed with sliding window.
    }
    \label{fig:sim_experiments}
\end{figure*}

\subsection{Experimental Setup}
\paragraph{HIL-RL Training Platform}

To enable fair and systematic comparisons, we introduce \newbench, a HIL-RL training suite with four simulated manipulation tasks. As shown in Fig.~\ref{fig:sim_env}, \newbench includes a long-horizon task, a standard pick-and-place task, an articulated-object manipulation task, and a deformable-object manipulation task. The first three tasks are implemented in MuJoCo, while \emph{Straighten-Rope} is adapted from AdaptiGraph~\cite{zhang2024adaptigraph}. Each episode terminates upon task success, a safety violation, or reaching the time limit, where safety violations include leaving the predefined workspace, exceeding velocity limits, and robot self-collision. We use the sparse reward setting described above, with a reward of $+10$ when task success is detected by predefined rules. A detailed comparison between \newbench and the existing HIL-RL benchmark Sirius~\cite{conf/rss/LiuNZBZ23} is provided in Appendix~\ref{supp:benchmark}. As shown in Fig.~\ref{fig:real_world_env}, we further evaluate all methods on two real-world tasks: \emph{Pick-Spoon}, which requires stable grasping under contact uncertainty, and \emph{Insert-USB}, which requires precise manipulation.

The simulated and real-world environments share a \emph{unified learning interface}, as shown in Fig.~\ref{fig:unified_interface}, which simplifies method development. For each real-world task, we additionally collect 20 demonstration trajectories, which are used to train a binary success classifier and are added to the intervention buffer to improve training efficiency for all methods. 

\paragraph{Human Interventions}
In \newbench, human-provided intervention actions are replaced with task-specific expert policies trained via RL, which provide consistent, reproducible corrections (see Appendix~\ref{supp:expert_model} for details). While these expert policies generate the intervention actions, human operators still decide \emph{when} to trigger interventions. Additional task details, like the maximum number of intervention rounds and the maximum episode length, are provided in Appendix~\ref{supp:task_details}.

\paragraph{Metrics}
We evaluate all methods using following metrics: 
\begin{enumerate}
    \item \textbf{Success Rate}: In real-world, it is computed over the most recent 10 training episodes. In simulation, it is measured over 100 autonomous trials at each evaluation interval.
    \item \textbf{Safety Violation Rate}: The faction of episodes terminated due to safety constraint violations. In some experiments, we report \textbf{Total Safety Violations} instead, which counts the cumulative safety violations during training. 
    \item \textbf{Episode Length}: It measures task-completion efficiency. For episodes completed with human intervention, we assign the maximum episode length to prevent human assistance from artificially improving this metric.
    \item \textbf{Intervention Ratio}: The proportion of human or expert intervention steps per training episode.

\end{enumerate}
In real-world evaluation, episodes with human intervention are counted as failures and assigned the maximum episode length, thereby excluding the effect of human assistance.

\paragraph{Baselines}
\ours is compared with 5 representative methods spanning HIL-RL, offline RL, interactive IL, and pessimistic exploration: \textbf{RLIF}~\cite{luo2024rlif}, \textbf{HIL-SERL}~\cite{luo2025precise}, \textbf{AWAC}~\cite{nair2020awac}, \textbf{IBRL}~\cite{DBLP:conf/rss/HuMS24}, and \textbf{Sirius}~\cite{conf/rss/LiuNZBZ23}, as discussed in Related Works. To ensure fair comparison, we modify some methods from their original implementations, as detailed in Appendix~\ref{supp:baseline}. Since \newbench provides no offline data, IBRL is included only for real-world tasks, where it is pretrained with 20 offline demonstrations. All methods share the same technical design from~\cite{luo2025precise}, including the intervention buffer, asynchronous actor-critic architecture, and pretrained encoder.

\subsection{Comparative Experiments}

\paragraph{Simulation Results}
As shown in Fig.~\ref{fig:sim_experiments}, we evaluate all methods on \newbench using success and safety violation rates to address \textbf{RQ1}. Results for intervention ratio and episode length are provided in Appendix~\ref{supp:exp}. \ours clearly outperforms the baselines on the two challenging tasks, \emph{Arrange-Boxes} and \emph{Straighten-Rope}. On \emph{Arrange-Boxes}, \ours achieves a 94\% success rate, improving over RLIF, HIL-SERL, AWAC, and Sirius by 44\%, 67\%, 63\%, and 75\%, respectively. Although both \ours and RLIF learn from intervention signals, the distributional Q-learning objective of \ours uses these signals more effectively. Its pessimistic exploration strategy also yields a policy with no safety violations on \emph{Arrange-Boxes}.

On the simpler \emph{Pick-Cube} and \emph{Open-Cabinet} tasks, \ours and AWAC reach similarly high success rates, while AWAC converges faster than the other baselines. This is likely because the expert provides reliable intervention actions, allowing AWAC to benefit from advantage-weighted behavior cloning and avoid out-of-distribution actions. On more complex tasks, however, \ours performs better by exploiting the information in intervention signals rather than relying mainly on imitation of corrective actions.

\begin{figure}[t]
    \centering
    \includegraphics[width=\linewidth]{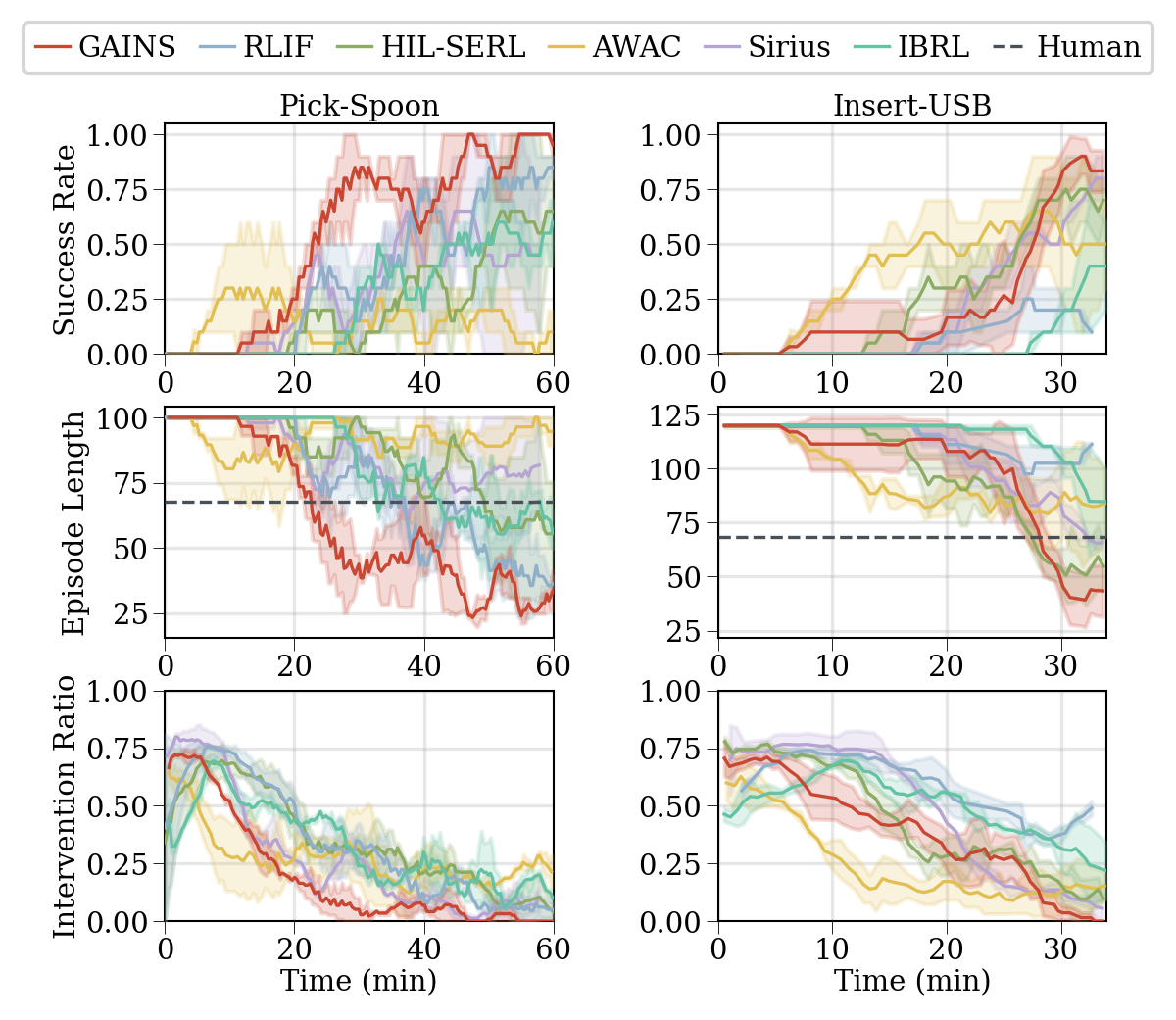}
    \caption{\textbf{Comparison Results in Real-World Tasks.} Results are averaged over 3 seeds.}

    \label{fig:real_world_exp}
\end{figure}

\paragraph{Real-world Results}
As shown in Fig.~\ref{fig:real_world_exp}, we evaluate on two real-world tasks. For episode length, we additionally report the average over 20 offline human demonstrations, denoted as ``Human''. Human interventions are provided through the bilateral teleoperation interface shown in Fig.~\ref{fig:bilteral-arm}. Overall, \ours reaches over 90\% success within 35--60 minutes and achieves shorter final episode lengths than the human demonstrations, indicating more efficient task completion.

AWAC shows relatively high initial success due to offline training but becomes unstable on \emph{Pick-Spoon}, which requires precise contact and gripper timing. This instability may result from inaccurate advantage estimates under sparse rewards, which can reinforce suboptimal grasping behaviors. Even with pretrained BC policy, IBRL does not show better initial performance than others. Perhaps due to 20 offline demonstrations provide insufficient coverage of diverse states during online training. RLIF performs poorly on \emph{Insert-USB}, likely due to delayed or inaccurate interventions caused by occlusion and the precision required for insertion.

In contrast, \ours remains robust to imperfect interventions by combining distributional Q-learning with pessimistic exploration, allowing it to use feedback effectively while reducing the impact of inconsistent signals.

\subsection{Robustness Evaluation}
\begin{figure}[t]
    \centering
    \includegraphics[width=\linewidth]{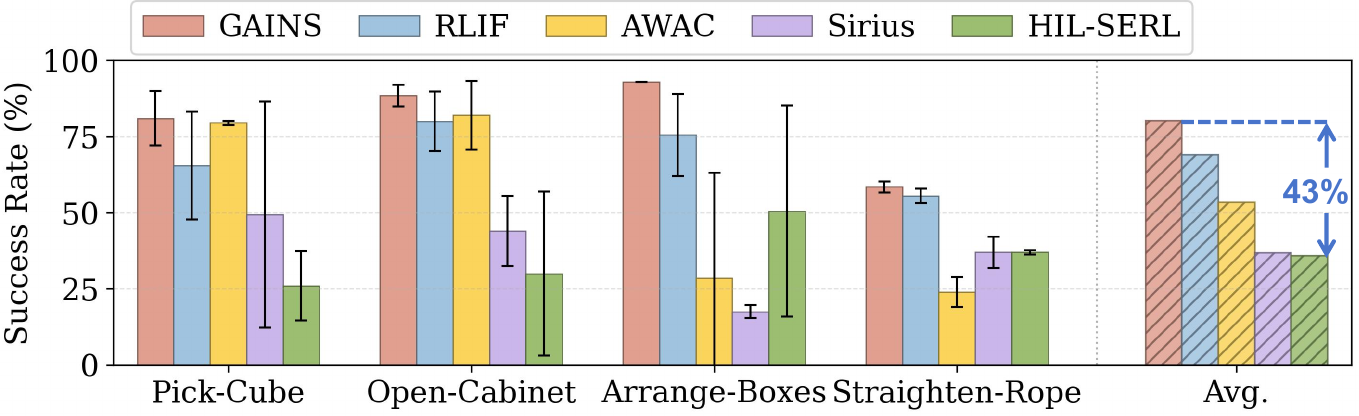}
    \caption{\textbf{Robustness Results in \newbench.} For each method, we evaluate the final checkpoints under 100 trials.}
    \label{fig:recovery}
\end{figure}

As shown in Fig.~\ref{fig:recovery}, we evaluate all methods under external disturbances, addressing \textbf{RQ2}. A recovery trajectory of \ours in robust evaluation is shown in Fig.~\ref{fig:robust_rope}. More details of the disturbance settings are provided in Appendix~\ref{supp:Disturbances}. Overall, \ours achieves an average recovery success rate of 80.3\%, outperforming HIL-SERL by 43\%. This result highlights the benefit of using information from intervention signals, which HIL-SERL does not explicitly exploit. AWAC performs competitively on \emph{Pick-Cube} and \emph{Open-Cabinet}, but performs much worse on \emph{Arrange-Boxes} and \emph{Straighten-Rope}. Since AWAC relies on advantage-weighted imitation over previously observed actions, it lacks an effective mechanism for handling these unseen recovery scenarios.

\begin{figure}[t]
    \centering
    \includegraphics[width=\linewidth]{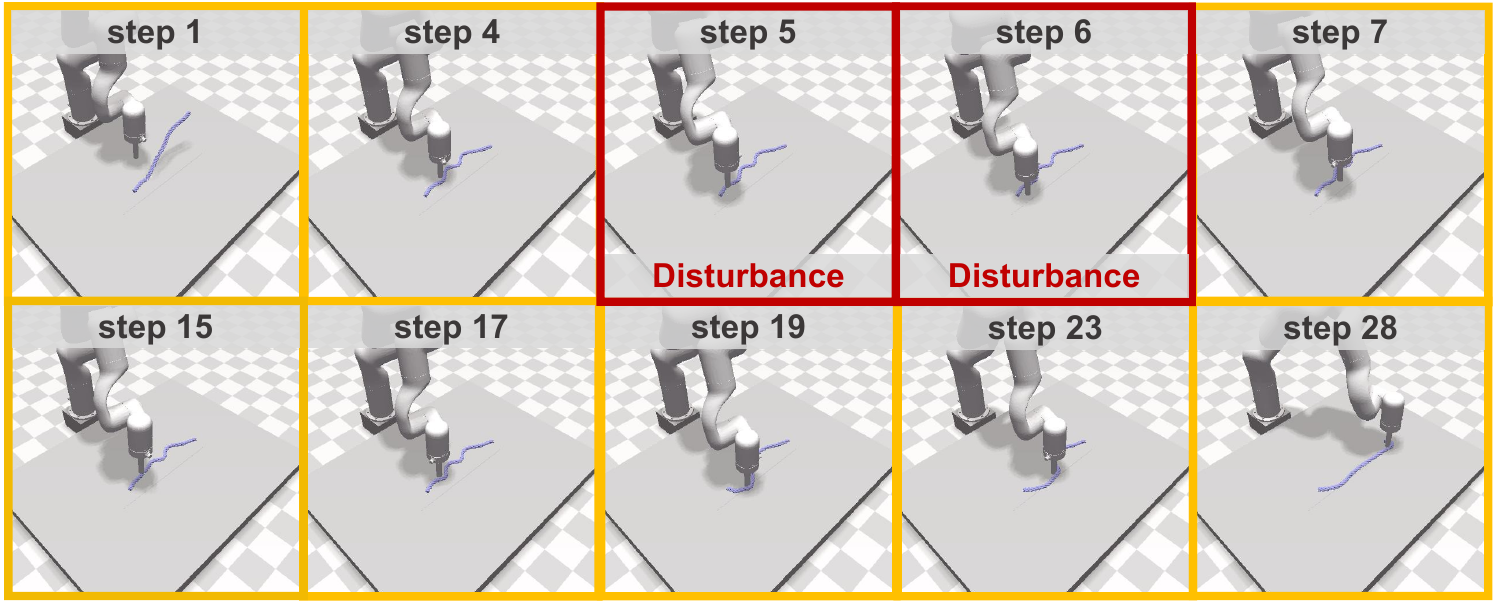}
    \caption{\textbf{Recovery Trajectory of \ours under External Disturbance.} More results can be found in Fig.~\ref{fig:robust_visual}}
    \label{fig:robust_rope}
\end{figure}

\begin{figure}[t]
    \centering
    \includegraphics[width=\linewidth]{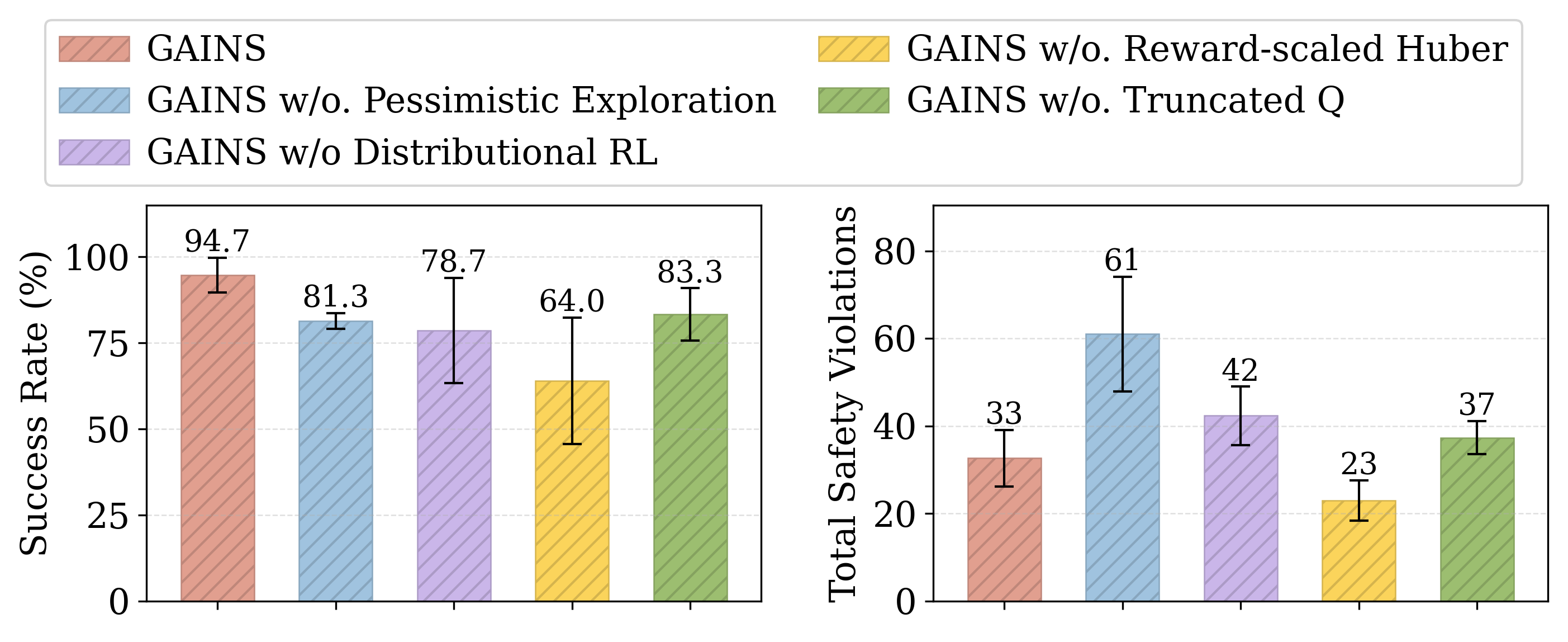}
    \caption{\textbf{Ablation Studies on Arrange-Boxes.} Each bar reports the mean performance over 3 random seeds. Total safety violations denote the cumulative number of safety violations throughout training.}
    \label{fig:ablation}
\end{figure}

\subsection{Ablation Studies}
To investigate the contribution of each component in \ours (\textbf{RQ3}), we conduct ablation studies on \emph{Arrange-Boxes}. As shown in Fig.~\ref{fig:ablation}, we evaluate four key design choices. First, removing pessimistic exploration and directly sampling actions from the learned action distribution (blue bar) reduces the success rate by 13.4\% and increases the total number of safety violations by an average of 28. This result highlights the importance of pessimistic exploration in reducing unsafe behaviors and the need for human intervention. Then, replacing distributional Q-value estimation with scalar Q-value estimation (purple bar) also causes a clear decrease in success rate, demonstrating the benefit of explicitly modeling the full distribution of mixed returns. Last, removing the reward-scaled Huber loss (yellow bar) reduces the success rate by 30.7\%, while also decreasing the number of safety violations. This is primarily because clipping the gradients at 1 slows the propagation of strong positive learning signals from successful outcomes, such as the $+10$ success reward, and consequently delays critic learning. The resulting policy behaves more conservatively, avoiding safety violations but also failing to make sufficient progress toward the task goal.

\subsection{Visualization of Value Distribution}

\begin{figure}[t]
    \centering
    \includegraphics[width=\linewidth]{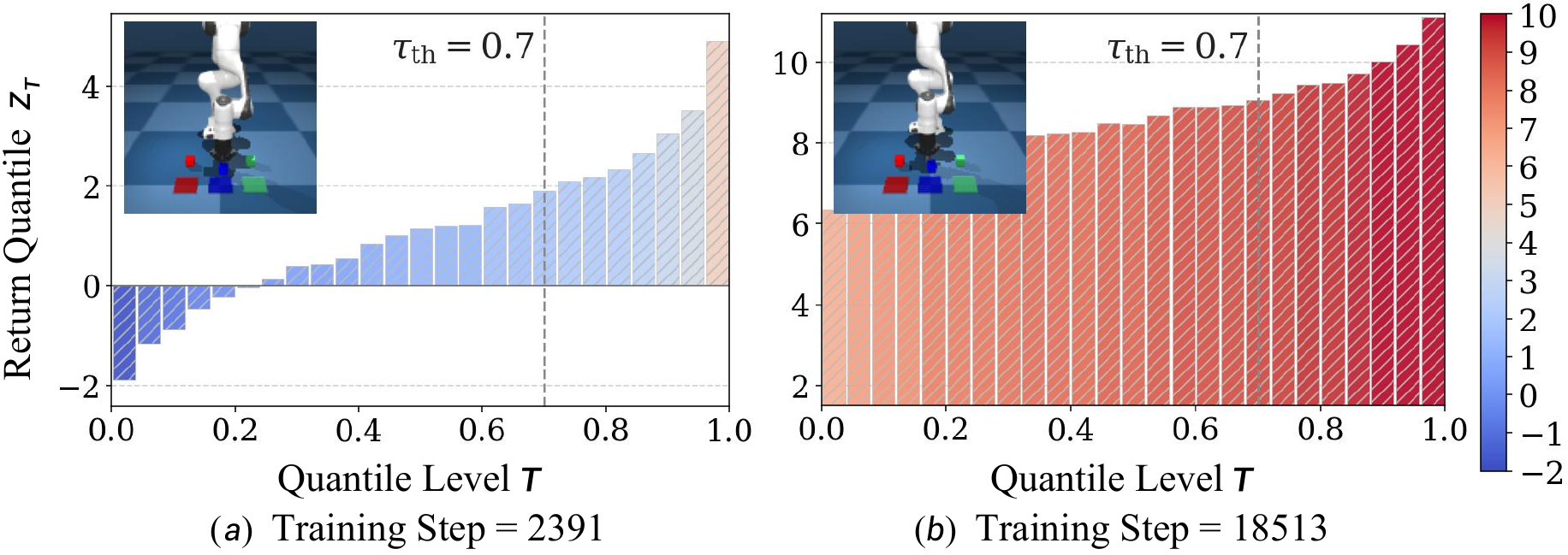}
    \caption{\textbf{Distributional Q-value Estimates at Different Stages of Training.}}
    \label{fig:value_distribution}
\end{figure}

To investigate whether tAAAI‘26he distributional critic captures uncertainty induced by human interventions, we visualize the learned return distributions at different training stages. As shown in the left panel of Fig.~\ref{fig:value_distribution}, during early training, pushing the blue boxes toward the goals may lead to task success, but often triggers human intervention or causes safety violations, such as pushing a box beyond the valid workspace. Consequently, \ours learns a high-variance return distribution, from which pessimistic exploration selects conservative actions using the quantile threshold $\tau_{\mathrm{th}}=0.7$. As training progresses, the lower-tail quantiles increase to above 6, indicating improved worst-case performance and more stable policy performance.

\section{Conclusion}
In this work, we propose \textbf{\ours}, a human-in-the-loop reinforcement learning (HIL-RL) framework for robust learning from inconsistent human interventions. Rather than explicitly estimating intervention uncertainty, \textbf{\ours} employs distributional Q-value estimation to model the heterogeneous return outcomes induced by human interventions and sparse rewards. Based on this representation, we introduce a reward-scaled quantile Huber loss for stable value learning and a pessimistic exploration strategy that selects actions according to their lower-tail returns. To enable systematic and fair evaluation, we develop a HIL training suite comprising four tasks. Experiments in simulation and on two real-world robotic systems show that \textbf{\ours} improves training efficiency over existing intervention-based methods while reducing unsafe exploration. On two real-world tasks, \textbf{\ours} also achieves shorter episode lengths than human demonstrations, indicating more efficient task execution.

\newpage

\bibliography{aaai2027}
\clearpage
\appendix

\setcounter{section}{0}
\setcounter{subsection}{0}
\setcounter{secnumdepth}{3}

\renewcommand{\thesection}{\Alph{section}}
\renewcommand{\thesubsection}{\thesection.\arabic{subsection}}
\renewcommand{\thesubsubsection}{\thesubsection.\arabic{subsubsection}}

\section{Benchmark Comparisons}\label{supp:benchmark}
As shown in Tab.~\ref{tab:sirius_hilrm_comparison}, we summarize the key differences between Sirius~\cite{conf/rss/LiuNZBZ23} and \newbench. A main advantage of HIL-RM is that it incorporates carefully pretrained RL experts to provide consistent intervention actions, which improves fairness across method comparisons and reduces human effort during intervention. HIL-RM also includes tasks with deformable and articulated objects, enriching the task diversity. Its current limitation is that it does not yet include dexterous manipulation tasks, such as assembly. To complement this limitation, we additionally report results on a real-world dexterous manipulation task, Insert-USB. We plan to include more tasks in future versions of \newbench.

\begin{table}[H]
\centering
\small
\setlength{\tabcolsep}{4pt}
\renewcommand{\arraystretch}{1.12}
\begin{tabularx}{\columnwidth}{
@{}
>{\raggedright\arraybackslash}X
>{\centering\arraybackslash}p{0.18\columnwidth}
>{\centering\arraybackslash}p{0.28\columnwidth}
@{}
}
\toprule
\textbf{Aspect} & \textbf{Sirius} & \textbf{HIL-RM} \\
\midrule
Intervention Action Source & Human & \makecell{Pretrained\\RL Expert} \\
Intervention Signal Source & Human & Human \\
Hardware Device & 3D Mouse & Keyboard \\
Task Numbers & 2 & 4 \\
\addlinespace[2pt]
Long-horizon Task & \checkmark & \checkmark \\
Deformable Manipulation & $\times$ & \checkmark \\
Articulated Object & $\times$ & \checkmark \\
Dexterous Manipulation & \checkmark & $\times$ \\
\bottomrule
\end{tabularx}
\caption{\textbf{Comparison between Sirius and HIL-RM.}}
\label{tab:sirius_hilrm_comparison}
\end{table}

\section{Environment Details in \newbench}\label{supp:sim_details}
In this section, we describe task-specific parameter settings, how the expert policies are trained, and how the disturbance environments are constructed in \newbench to facilitate evaluation during peer review. We will open-source the environments and all trained models to promote research in HIL-RL and embodied AI more broadly.

\subsection{Task Settings}\label{supp:task_details}
The performance of HIL methods can depend on the amount of human intervention. To prevent excessive intervention and ensure a fair and practical evaluation, we limit the number of intervention decisions during training. Specifically, the human operator may initiate at most $N_{\text{interv}}$ continuous intervention segments per episode.

For robustness evaluation, we control the disturbance intensity using $N_{\text{noise}}$, which denotes the number of consecutive disturbance steps. A larger $N_{\text{noise}}$ causes greater deviation from the training-state distribution and therefore presents a more challenging robustness test. For each task, Tab.~\ref{tab:pick_cube} to Tab.~\ref{tab:rope} report $N_{\text{interv}}$, $N_{\text{noise}}$, the episode length, and the state and action spaces.

\begin{table}[htp]
\centering
\small
\setlength{\tabcolsep}{8pt}
\renewcommand{\arraystretch}{1.12}
\begin{tabular}{lc}
\toprule
\textbf{Parameter} & \textbf{Value}\\
\midrule
Episode Length & 50  \\
$N_{\text{interv}}$ & 5  \\
$N_{\text{noise}}$  & 5   \\
State Space &
$\mathbf{q}\in\mathbb{R}^7$, $\dot{\mathbf{q}}\in\mathbb{R}^7$, $\mathbf{p}_{tcp}\in\mathbb{R}^3$, $g\in\mathbb{R}$\\
Action Space &
$\mathbf{p}\in\mathbb{R}^3$,$\mathbf{r}\in\mathbb{R}^3$,  $g\in\mathbb{R}$ \\
\bottomrule
\end{tabular}
\caption{\textbf{Pick-Cube} $\mathbf{q}$ and $\dot{\mathbf{q}}$ denote joint positions and velocities,
$\mathbf{p}_{tcp}$ denotes the TCP position,
$\mathbf{p}$ and $\mathbf{r}$ denote Cartesian position and orientation commands,
and $g$ denote the gripper state/command.}
\label{tab:pick_cube}
\end{table}

\begin{table}[htp]
\centering
\small
\setlength{\tabcolsep}{8pt}
\renewcommand{\arraystretch}{1.12}
\begin{tabular}{lc}
\toprule
\textbf{Parameter} & \textbf{Value}\\
\midrule
Episode Length & 50  \\
$N_{\text{interv}}$ & 5  \\
$N_{\text{noise}}$  & 5   \\
State Space &
$\mathbf{p}_{tcp}\in\mathbb{R}^3$,$\mathbf{q}_{tcp}\in\mathbb{R}^4$,$g\in\mathbb{R}$,$d_{door}\in\mathbb{R}^3$ \\
Action Space & $\mathbf{p}\in\mathbb{R}^3$,$\mathbf{r}\in\mathbb{R}^3$,$g\in\mathbb{R}$ \\
\bottomrule
\end{tabular}
\caption{\textbf{Open-Cabinet.} $\mathbf{q}_{tcp}$ denote the orientation, and $d_{door}\in\mathbb{R}^3$ denotes the cabinet door opening state replicated across three dimensions.}
\label{tab:open_cabinet}
\end{table}

\begin{table}[htp]
\centering
\small
\setlength{\tabcolsep}{8pt}
\renewcommand{\arraystretch}{1.12}
\begin{tabular}{lc}
\toprule
\textbf{Parameter} & \textbf{Value}\\
\midrule
Episode Length & 200  \\
$N_{\text{interv}}$ & 15  \\
$N_{\text{noise}}$  & 10   \\
State Space & $\mathbf{p}_{tcp}\in\mathbb{R}^{6}$,$\mathbf{p}_{box}\in\mathbb{R}^{18}$ \\
Action Space & $\mathbf{p}\in\mathbb{R}^{3}$ \\
\bottomrule
\end{tabular}
\caption{\textbf{Arrange-Boxes.} $\mathbf{p}_{tcp}\in\mathbb{R}^{6}$ denotes the TCP position obtained by duplicating the original 3D TCP coordinates. $\mathbf{p}_{box}\in\mathbb{R}^{18}$ denotes the positions of three boxes, where the 3D position of each box is duplicated to form a 6D representation.}
\label{tab:arrange_box}
\end{table}

\begin{table}[htp]
\centering
\small
\setlength{\tabcolsep}{8pt}
\renewcommand{\arraystretch}{1.12}
\begin{tabular}{lc}
\toprule
\textbf{Parameter} & \textbf{Value}\\
\midrule
Episode Length & 70  \\
$N_{\text{interv}}$ & 10  \\
$N_{\text{noise}}$  & 2   \\
State Space & $\mathbf{p}_{tcp}\in\mathbb{R}^3$,$\mathbf{q}_{tcp}\in\mathbb{R}^4$\\
Action Space & $\mathbf{p}\in\mathbb{R}^{3}$ \\
\bottomrule
\end{tabular}
\caption{\textbf{Straighten-Rope.} For the Rope task, the vertical action $z$ is discretized to $\{0,1\}$ and selected by a discrete critic, while $x$ and $y$ remain continuous.}
\label{tab:rope}
\end{table}

\subsection{Built-in Expert Model}\label{supp:expert_model}
We provide a built-in expert policy for each task in \newbench. Each expert is trained using a task-specific RL algorithm to generate effective recovery actions when a human intervention signal is issued. For \emph{Pick-Cube} and \emph{Open-Cabinet} tasks, expert policies are trained using SAC. For \emph{Arrange-Boxes} and \emph{Straighten-Rope}, SAC exhibited slow and unstable convergence due to the long-horizon nature of the tasks and the complexity of flexible-object manipulation. Therefore, expert policies for these tasks are trained using \ours . We employ dense reward functions to facilitate exploration and accelerate convergence as shown below.

To improve the recovery capability of expert policies, we adopt a two-stage training procedure. The expert is first trained under the standard environment setting. Once the success rate reaches approximately 50\%, action noise is injected during training. This exposes the policy to off-nominal states and encourages it to recover from execution errors, thereby producing more effective intervention actions.

\paragraph{Pick-Cube.}
$r_t=0.3d_{\text{reach}}+0.7d_{\text{lift}}$,
where $d_{\text{reach}}=\exp(-20||p_{\text{gripper}}-p_{\text{cube}}||_2)$ denotes the normalized reaching reward based on the gripper-cube distance, and $d_{\text{lift}}$ is the normalized lifting height.

\paragraph{Open-Cabinet.}
$r_t=0.3d_{\text{handle}}+0.7d_{\text{door}}/d_{\max}$,
where $d_{\text{handle}}=\exp(-5||p_{\text{gripper}}-p_{\text{handle}}||_2)$ denotes the normalized reaching reward based on the gripper-handle distance, and $d_{\text{door}}$ denotes the current cabinet door displacement with $d_{\max}=0.22$ being the maximum opening distance. The ratio $d_{\text{door}}/d_{\max}$ normalizes the opening progress to $[0,1]$.

\paragraph{Arrange-Boxes.}
$r_t=\sum_i (d_i^{(t-1)}-d_i^{(t)})$, where $d_i$ is the distance between the $i$-th box and its target and a penalty of $-0.1$ is applied when no progress is made.

\paragraph{Straighten-Rope.}
$r_t=\mathrm{clip}(50(d_{t-1}-d_t),0,1)$, where $d_t$ denotes the average particle deviation from the target line.

\begin{figure}[ht]
  \centering
    \begin{subfigure}[htp]{\columnwidth}
    \centering
    \includegraphics[width=\linewidth]{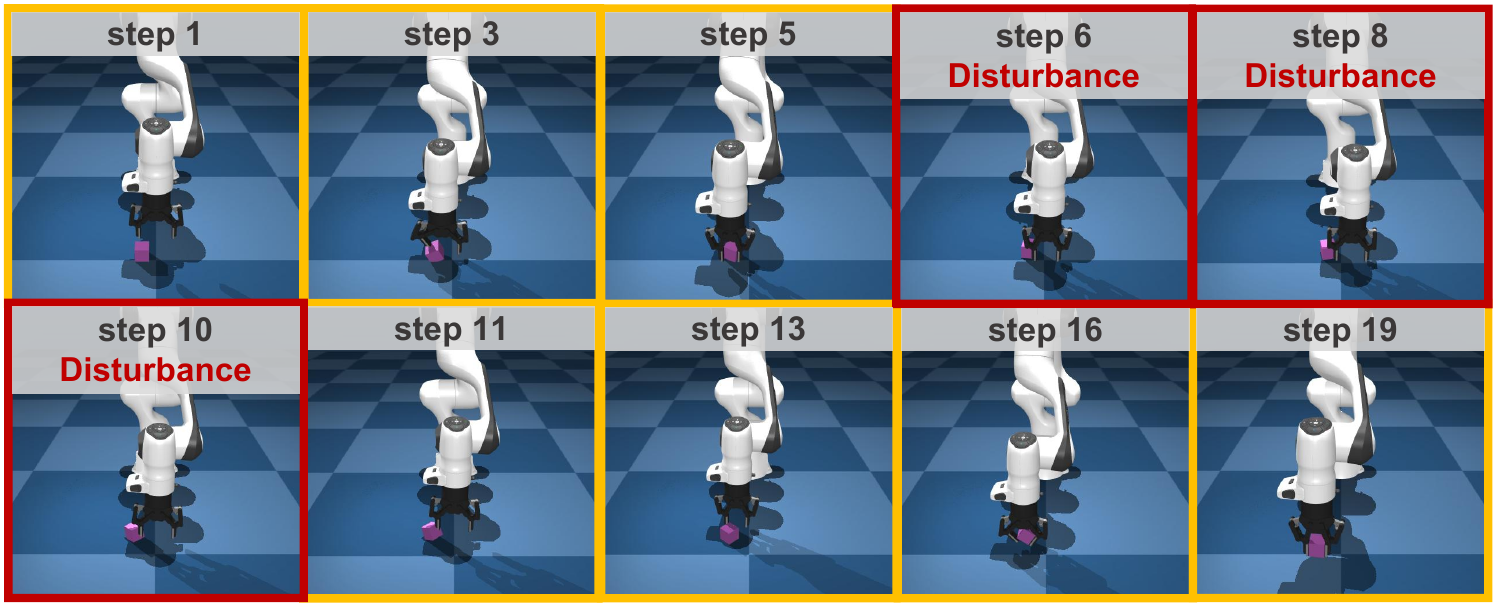}
    \caption{Pick-cube.}
    \label{fig:robust_pick_cube}
  \end{subfigure}
  \begin{subfigure}[htp]{\columnwidth}
    \centering
    \includegraphics[width=\linewidth]{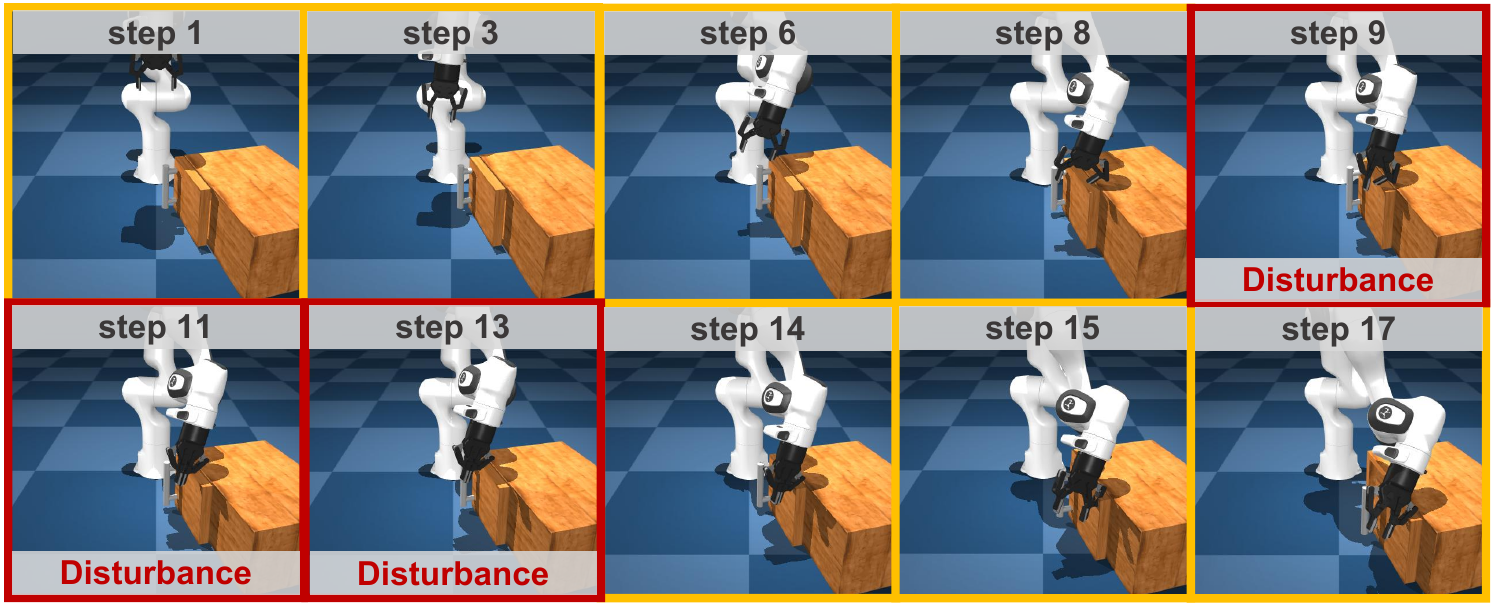}
    \caption{Open-Cabinet.}
    \label{fig:robust_cabinet}
  \end{subfigure}

  \begin{subfigure}[htp]{\columnwidth}
    \centering
    \includegraphics[width=\linewidth]{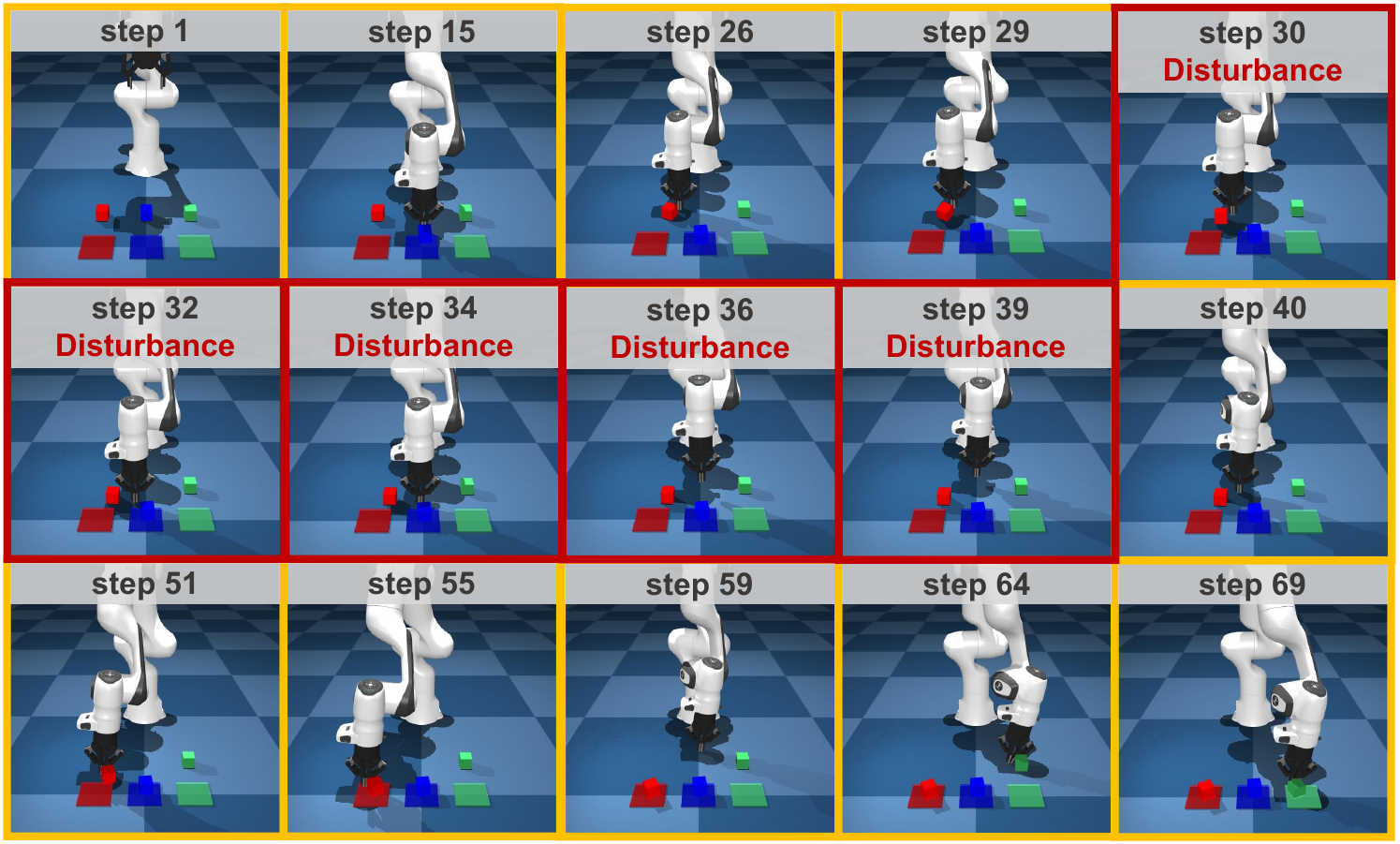}
    \caption{Arrange-boxes.}
    \label{fig:robust_arrange}
  \end{subfigure}
  
  \caption{
  \textbf{Robust Environment of \newbench.} 
  }
  \label{fig:robust_visual}
\end{figure}

\subsection{Rule-based External Disturbances}\label{supp:Disturbances}

In \emph{Pick-Cube} and \emph{Arrange-Boxes}, disturbances are injected at a randomly selected time step in each episode, as shown in Figs.~\ref{fig:robust_pick_cube} and~\ref{fig:robust_arrange}. After being triggered, the disturbance is implemented through temporary action corruption, where Gaussian noise is added to the robot actions for a fixed number of consecutive steps. This causes the robot to deviate from its nominal trajectory and requires the policy to exhibit recovery behavior.

In \emph{Open-Cabinet}, disturbances are injected after the end-effector contacts with the cabinet, as shown in Fig.~\ref{fig:robust_cabinet}. The disturbance is implemented in one of two ways with equal probability: either Gaussian action noise is applied, or the end-effector is forced to move away from the handle. Both cases create a contact failure that requires the policy to recover before continuing the task.

In \emph{Straighten-Rope}, disturbances are injected once the end-effector contacts the rope, as shown in Fig.~\ref{fig:robust_rope}. The disturbance is implemented by amplifying the commanded motion by a factor of two during the disturbance phase. Additionally, in 50\% of the episodes, the vertical motion command is inverted, causing lifting motions to become lowering motions and vice versa. These disturbances intentionally disrupt the interaction between the end-effector and the rope, requiring the policy to recover and continue the straightening process.

\section{Infrastructure Details}
\subsection{Homogeneous-arm Setup}\label{supp:bilteral-arm}
\begin{figure}[htp]
    \centering
    \includegraphics[width=0.8\columnwidth]{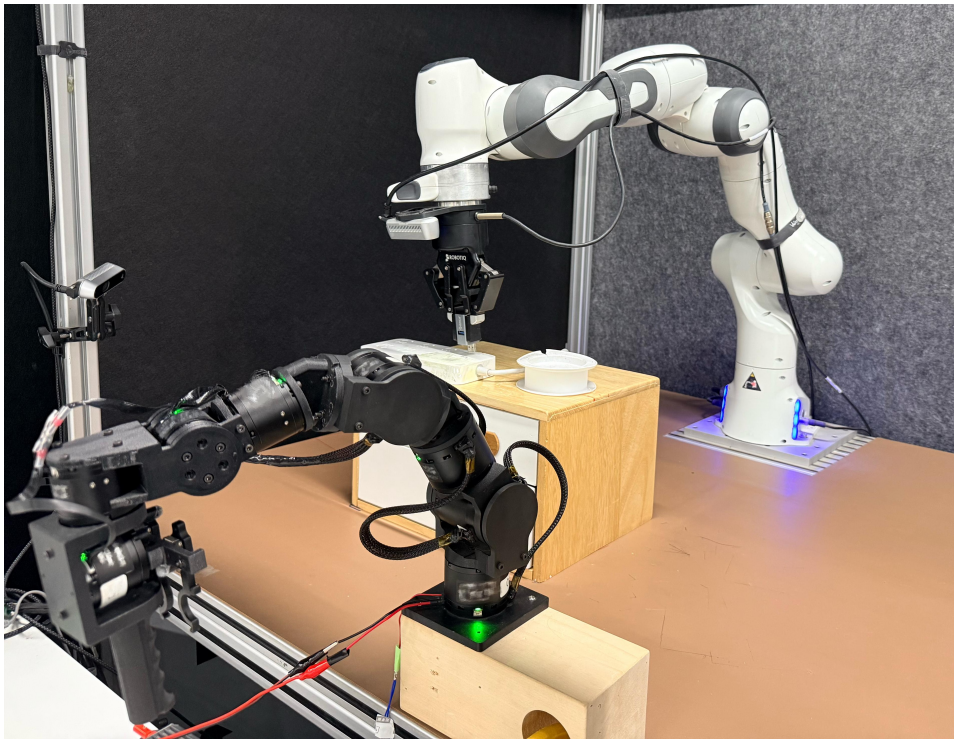}
    \caption{\textbf{Homogeneous bilateral manipulator setup in Franka Emika.} }
    \label{fig:bilteral-arm}
\end{figure}

Fig. ~\ref{fig:bilteral-arm} shows the homogeneous bilateral manipulator system used in our real-world experiments. During intervention, the master manipulator is directly controlled by a human operator, while the puppy manipulator executes the corresponding actions in the task workspace. Since the two manipulators share the same kinematic structure, an intuitive pose-to-pose mapping can be established. 

\subsection{Computing Infrastructure}

All experiments were conducted on workstations running Ubuntu 22.04 LTS. Simulation experiments, including policy training and inference, were performed on a workstation equipped with a single NVIDIA GeForce RTX 4090 GPU with 24,GB of memory, an AMD Ryzen 9 9950X processor with 16 physical cores and 32 threads, and 60,GB of system memory.

For real-world experiments, the actor and learner were deployed on separate workstations. The actor workstation was equipped with an NVIDIA GeForce RTX 4090 GPU with 24,GB of memory and an AMD Ryzen 9 9950X processor with 16 cores and 32 threads. The learner workstation used an NVIDIA RTX A6000 GPU with 48,GB of memory and an Intel Core i9-13900K processor with 32 logical CPUs.

\subsection{System Implementation}
\begin{figure}[H]
    \centering
    \includegraphics[width=\columnwidth]{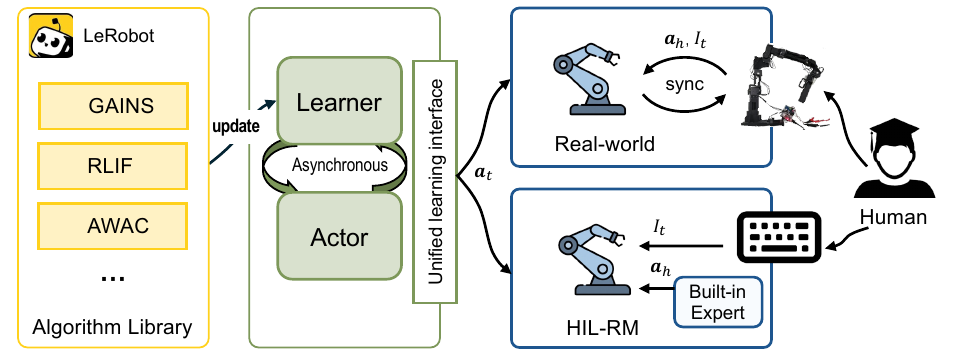}
    \caption{\textbf{System Structure in \ours.} }
    \label{fig:unified_interface}
\end{figure}

\ours provides a unified real-sim HIL-RL training system, as shown in Fig.~\ref{fig:unified_interface}. Human operators can intervene through different devices in either simulation or real-world tasks, while the training pipeline is unified under an asynchronous actor--critic learning framework built on LeRobot~\cite{lerobot}. This modular design decouples teleoperation, environment interaction, data collection, and policy learning, making the system easy to extend to new robots, methods, tasks, and intervention interfaces. Our implementation uses Python 3.10.19, PyTorch 2.7.1 with CUDA 12.4, and LeRobot 0.4.2.

\begin{figure*}[t]
    \centering
    \includegraphics[width=\linewidth]{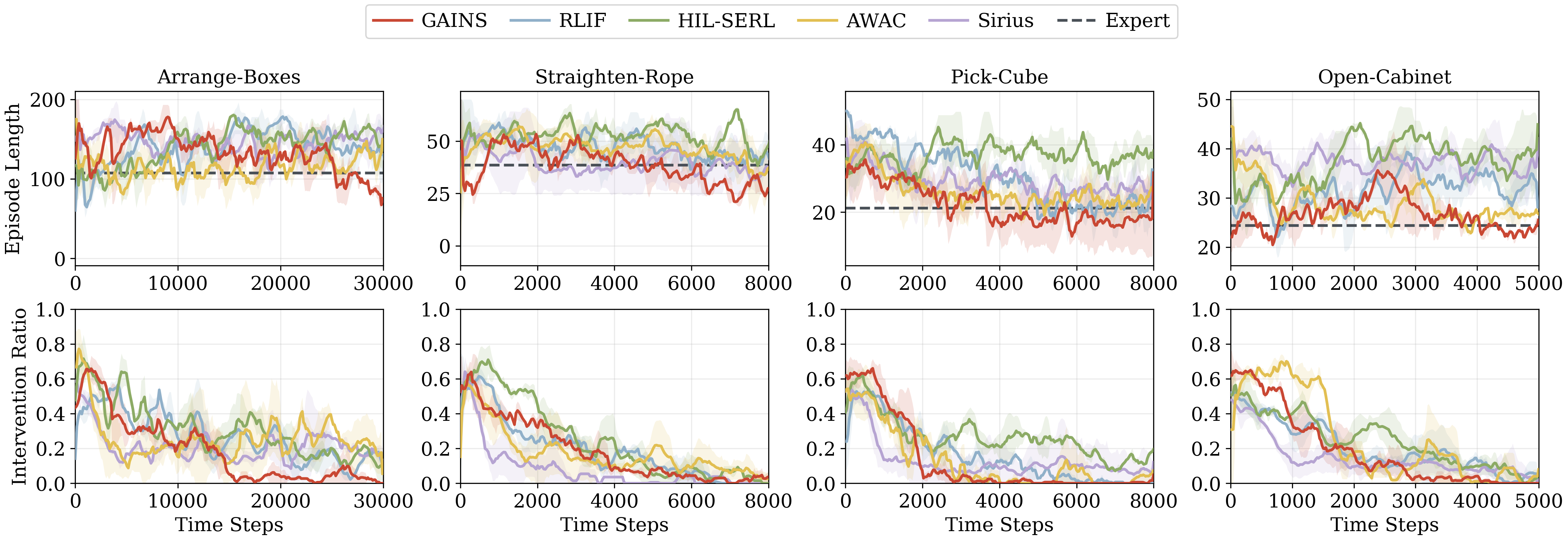}
    \caption{\textbf{Comparison Results on Episode Length and Intervention Ratio in \newbench.} Results are averaged over three seeds, with shaded regions indicating the standard deviation. Curves are smoothed with sliding window.
    }
    \label{fig:sim_interv_episode}
\end{figure*}

\section{Implementation Details}\label{supp:Hyperparameters}
\subsection{\ours Implementation}\label{supp:gains_details}
In \ours, the actor is parameterized by a two-layer MLP with 256 hidden units per layer, SiLU activations, and layer normalization. The actor and critic use separate encoder instances. For visual observations, following HIL-SERL~\cite{luo2025precise}, we employ a frozen ResNet-10 backbone. Spatial feature maps are aggregated using learned spatial embeddings and projected into a 256-dimensional latent representation for each camera view. Tab.~\ref{tab:method_details} summarizes the key hyperparameters used in \ours.

\begin{figure}[t]
    \centering
    \includegraphics[width=0.7\columnwidth]{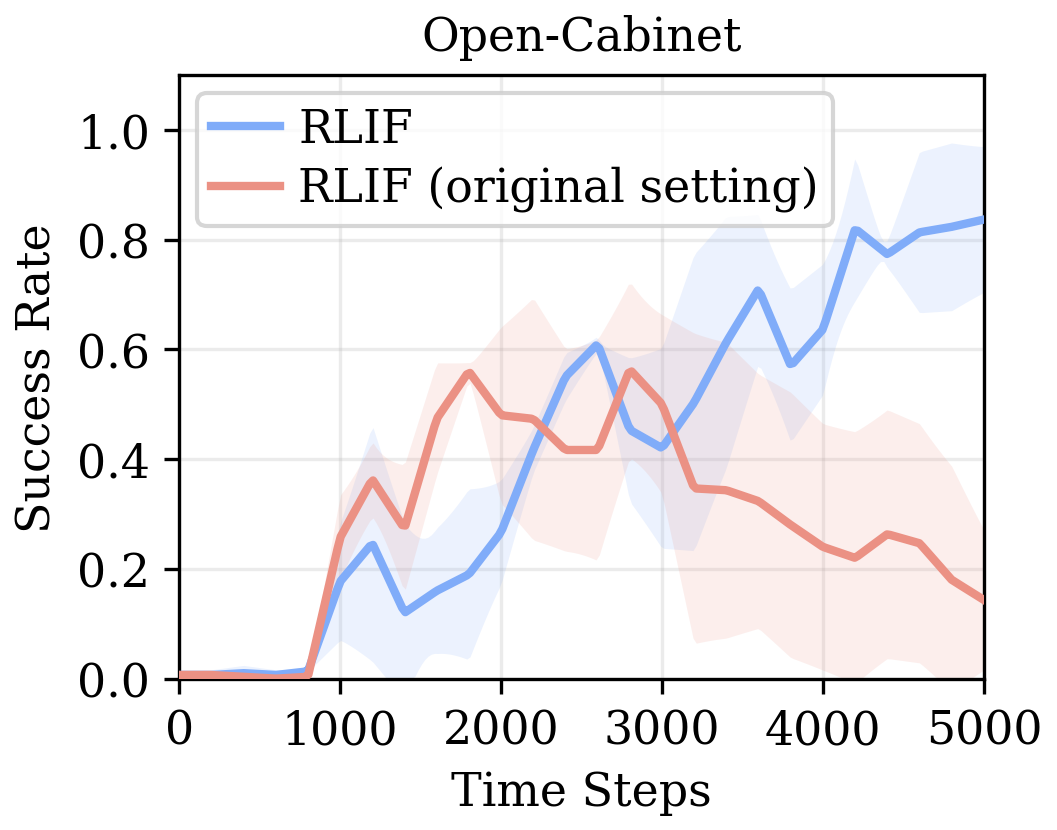}
    \caption{\textbf{RLIF performance under our implementation (blue) and the original configuration (red) on \emph{Open-Cabinet}.}}
    \label{fig:rlif-compare}
\end{figure}

\subsection{Baselines Implementation}\label{supp:baseline}
We compare \ours against the following representative baselines for online training:
\begin{itemize}

\item \textbf{RLIF}~\cite{luo2024rlif} is an RL method that uses only intervention signals as feedback. For a fair comparison, we additionally provide success rewards and safety penalties. As shown in Fig.~\ref{fig:rlif-compare}, the original RLIF setting (red) performs similarly to our setting (blue) during the first 3k steps but then degrades rapidly, as it learns primarily to avoid interventions without explicit guidance toward the task goal.

\item \textbf{HIL-SERL}~\cite{luo2025precise} is a real-world method that integrates demonstrations and human corrections through sample-efficient RL and system-level optimizations. 

\item \textbf{AWAC}~\cite{nair2020awac} is an offline RL baseline that performs advantage-weighted policy updates.

\item \textbf{Sirius}~\cite{conf/rss/LiuNZBZ23} is an interactive IL method that uses human intervention signals to estimate human trust and reweight training samples accordingly. For a fair comparison, we adapt Sirius to our setting by enabling frequent online updates and using the same model architecture as ours.

\item \textbf{IBRL}~\cite{DBLP:conf/rss/HuMS24} is a sample-efficient RL method that uses pretrained IL policy to propose alternative actions for both online exploration and target value bootstrapping. Since \newbench provides no offline data, we only include IBRL for real-world tasks, which are pretrained using 20 offline demonstrations.
\end{itemize}
All methods share the same technical designs implemented in~\cite{luo2025precise}, including the additional intervention buffer, asynchronous actor-critic architecture, and pretrained encoder.

\begin{table}[t]
\centering
\small
\setlength{\tabcolsep}{8pt}
\renewcommand{\arraystretch}{1.12}
\begin{tabular}{p{5.5cm}c}
\toprule
\textbf{Parameter} & \textbf{Value}\\
\midrule
Truncated Q Drop Dim $b$                                                                          & 2 \\
Critic Number $M$                                                                                 & 2 \\
Quantile Bins $K$                                                                                 & 25 \\
Huber Threshold $\kappa$                                                                          & 10 \\
Pessimistic Exploration Threshold $\tau_{\mathrm{th}}$ (Pick-Cube, Straighten-Rope)              & 0.5 \\
Pessimistic Exploration Threshold $\tau_{\mathrm{th}}$ (Arrange-Boxes, Open-Cabinet)             & 0.7 \\
Learning rate                                                                                    & 3e-4 \\
Discount factor                                                                                  & 0.97 \\
Optimizer                                                                                        & Adam \\
\bottomrule
\end{tabular}
\caption{\textbf{Key Hyperparameters of \ours.}}
\label{tab:method_details}
\end{table}

\section{Additional Results}\label{supp:exp}
\subsection{Additional Comparison Results}
As shown in Fig.~\ref{fig:sim_interv_episode}, we report the episode length and intervention ratio across all methods during training. We observe the intervention ratio consistently decreases for all methods as training proceeds, indicating reduced reliance on human interventions and improved task performance. Notably, \ours converges to the lowest intervention ratio in later stages across all tasks, suggesting fully autonomous task completion. This is mainly because \ours effectively distills information from early-stage interventions into the Q function through distributional Q-learning.

For episode length, we observe a clear task-dependent separation due to different expert level. HIL-SERL shows only marginal reductions across all tasks, resulting in consistently longer episodes, which can be attributed to its limited ability to exploit the information in the replayed intervention data for accelerating recovery from suboptimal states. In contrast, \ours achieves significantly shorter episodes, particularly on \emph{Arrange-Boxes} and \emph{Straighten-Rope}, even surpassing expert-level efficiency. This improvement reflects more efficient policy execution enabled by better utilization of intervention signals via distributional Q-learning and pessimistic exploration, where the learned value distribution improves credit assignment for corrective actions and promotes faster convergence to stable and efficient trajectories, thereby reducing redundant steps during task execution.

\subsection{Investigation on Quantile Threshold}
\begin{figure}[t]
  \centering
    \begin{subfigure}[t]{\columnwidth}
    \centering
    \includegraphics[width=\linewidth]{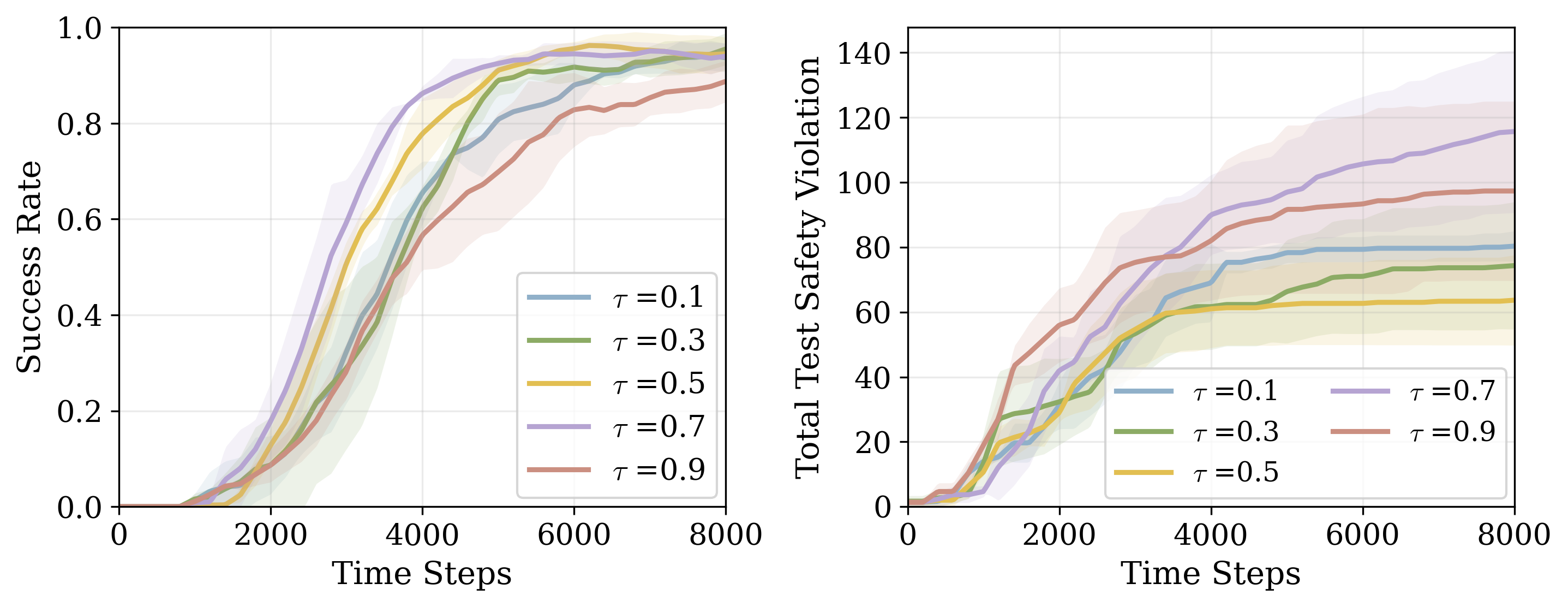}
    \caption{Pick-cube.}
    \label{fig:tau_top}
  \end{subfigure}
  \begin{subfigure}[t]{\columnwidth}
    \centering
    \includegraphics[width=\linewidth]{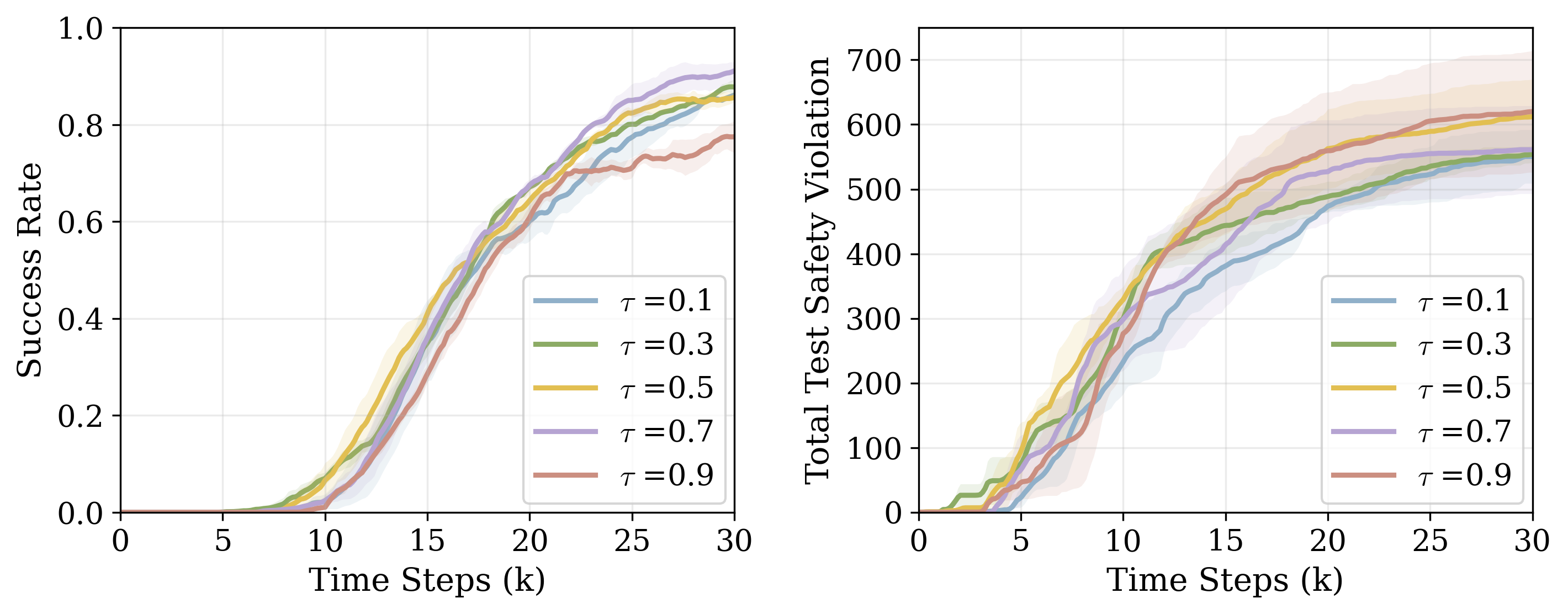}
    \caption{Arrange-boxes.}
    \label{fig:tau_bottom}
  \end{subfigure}
  \caption{
  \textbf{Investigation of Quantile Threshold $\tau_\textrm{th}$.} Results are averaged over 3 seeds.
  }
\end{figure}

Recall that in Fig.~\ref{fig:overview}, at each time step, we sample a set of candidate actions and select the one with the maximum accumulated Q estimate under a predefined quantile threshold $\tau_\textrm{th}$. In this paragraph, we vary $\tau_\textrm{th}$ from 0.1 to 0.9 to investigate how \ours responds to this key parameter. As shown in Fig.~\ref{fig:tau_top}, for \emph{Pick-cube} task, setting $\tau_\textrm{th}$ between 0.1 and 0.7 leads to similarly high converged success rates, whereas $\tau_\textrm{th}=0.9$ results in noticeably less efficient convergence. Similar results can be observed in the \emph{Arrange-Boxes} task in Fig.~\ref{fig:tau_bottom}, where only \ours with $\tau_\textrm{th}=0.9$ converges to a success rate below 80\%.

This is because we adopt a large success reward of +10 under sparse reward settings, following previous works~\cite{luo2025precise,chen2025conrft}. A large $\tau_\textrm{th}$ makes the Q estimate more optimistic and may underemphasize safety violations and intervention penalties when high success returns dominate the value distribution. As a result, the policy is more likely to select risky or unreliable actions, leading to less efficient training.

An overly conservative exploration strategy can also degrade convergence by slowing down policy improvement. For example, as shown in Fig.~\ref{fig:tau_top}, setting $\tau_\textrm{th}=0.1$ leads to clearly slower convergence, achieving a success rate 10\% lower than that of $\tau_\textrm{th}=0.7$ after 4k training steps. This indicates that task rewards still play an important role in guiding efficient learning, and excessive pessimism may prevent the policy from exploiting high-return actions. Overall, setting $\tau_\textrm{th}=0.5$ for \emph{Pick-Cube} and $\tau_\textrm{th}=0.7$ for \emph{Arrange-Boxes} achieves efficient training while maintaining policy safety.

\end{document}